\documentclass[10pt,twocolumn,letterpaper]{article}

\usepackage[pagenumbers]{cvpr} 

\definecolor{cvprblue}{rgb}{0.21,0.49,0.74}
\usepackage[pagebackref,breaklinks,colorlinks,allcolors=cvprblue]{hyperref}
\usepackage{placeins} 
\usepackage[accsupp]{axessibility}  

\usepackage{multirow} 

\def\paperID{10137} 
\def\confName{CVPR}
\def\confYear{2026}

\title{\textbf{OpenMarcie}: Dataset for Multimodal Action Recognition in Industrial Environments}

\author{
Hymalai Bello$^{1 \dagger}$ \quad
Lala Ray$^{1 \dagger}$ \quad
Joanna Sorysz$^{1,2}$ \quad
Sungho Suh$^{3*}$ \quad
Paul Lukowicz$^{1,2}$ \\
$^{1}$DFKI Kaiserslautern \quad
$^{2}$RPTU Kaiserslautern-Landau \quad
$^{3}$Korea University\\
{\tt\small
\{Hymalai.Bello, Lala\_Shakti\_Swarup.Ray, Joanna.Sorysz, Paul.Lukowicz\}@dfki.de} \\
{\tt\small sungho\_suh@korea.ac.kr}\\
{\small
$^{\dagger}$Equal contribution \quad
$^{*}$Corresponding author}
}

\begin{document}
\maketitle
\begin{abstract}
Smart factories use advanced technologies to optimize production and increase efficiency. 
To this end, the recognition of worker activity allows for accurate quantification of performance metrics, improving efficiency holistically while contributing to worker safety.
OpenMarcie is, to the best of our knowledge, the biggest multimodal dataset designed for human action monitoring in manufacturing environments. It includes data from wearables sensing modalities and cameras distributed in the surroundings. 
The dataset is structured around two experimental settings, involving a total of 36 participants.
In the first setting, twelve participants perform a bicycle assembly and disassembly task under semi-realistic conditions without a fixed protocol, promoting divergent and goal-oriented problem-solving.
The second experiment involves twenty-five volunteers (24 valid data) engaged in a 3D printer assembly task, with the 3D printer manufacturer's instructions provided to guide the volunteers in acquiring procedural knowledge.
This setting also includes sequential collaborative assembly, where participants assess and correct each other's progress, reflecting real-world manufacturing dynamics.
OpenMarcie includes over 37 hours of egocentric and exocentric, multimodal, and multipositional data, featuring eight distinct data types and more than 200 independent information channels. 
The dataset is benchmarked across three human activity recognition tasks: activity classification, open vocabulary captioning, and cross-modal alignment. 
The dataset and code are available at \href{https://kaggle.com/datasets/4942046dbb2608f5a0c3e76c9bd53c6b9b2784d738da7db37f7739b89c2c7f95}{OpenMarcie}. 
\end{abstract}    
\section{Introduction}
\label{sec:intro}
\begin{figure*}[]
    \centering
    \includegraphics[width=0.95\linewidth]{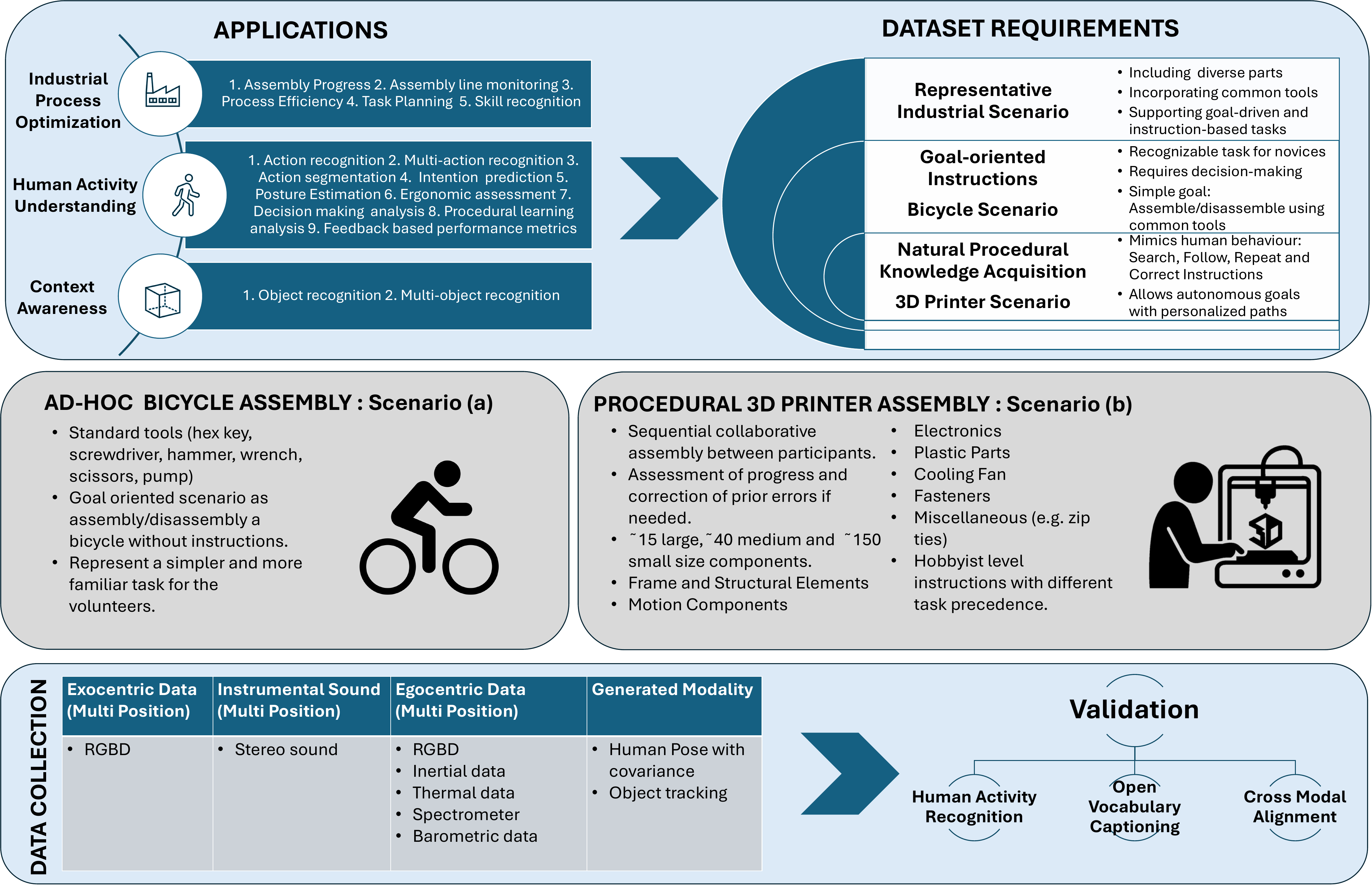}
    \caption{OpenMarcie is an open dataset for multimodal action recognition in industrial environments.
    \textbf{Top} Its applications include industrial process optimization, human activity understanding, and context awareness. 
    \textbf{Middle} two assembly scenarios represent ad-hoc goal-oriented action and a procedural scenario to promote natural knowledge acquisition. 
    \textbf{Bottom} multimodal data is collected with more than 200 independent channels, and a validation with three different benchmarks is done.
 }
    \label{fig:Marcie1}
    \vspace{-5mm}
\end{figure*}

\begin{table*}[t!]
\caption{Comparative table of human action recognition datasets in the industrial domain. Multi-actions indicates whether the dataset supports concurrent action labeling (e.g., walking while carrying), based on overlapping verb–object–tool annotations. Only OpenMarcie unifies \textbf{wearables, egocentric + exocentric multiview video, multi-action labels, and full industrial coverage}.}

\centering
\footnotesize
\setlength{\tabcolsep}{6pt}
\renewcommand{\arraystretch}{0.8}
\resizebox{\textwidth}{!}{
\begin{tabular}{lccccccc}
\hline
Dataset & Industrial & Wearables & Ego & Exo & Multiview & Multi-Act. & Scale / Focus \\
\hline
InHARD \citep{dallel2020inhard}       & \checkmark & \checkmark & x & \checkmark & \checkmark & x & HRC assembly, RGB+IMU \\
LARa \citep{niemann2020lara}        & \checkmark & \checkmark & x & \checkmark & x & x & Logistics, IMU+mocap \\
OpenPack \citep{yoshimura2024openpack}    & \checkmark & \checkmark & x & \checkmark & x & x & Logistics, 50h+ IMU/IoT \\
Assembly101 \citep{sener2022assembly101} & \checkmark & x & \checkmark & \checkmark & \checkmark & x & Vision-only, procedural assembly \\
IKEA-ASM \citep{ben2021ikea}    & \checkmark & x & x & \checkmark & \checkmark & x & Exo RGB-D, furniture assembly \\
HA4M \citep{cicirelli2022ha4m}       & \checkmark & x & x & \checkmark & x & x & Exo RGB-D+IR+pose \\
HA-ViD \citep{zheng2023ha}       & \checkmark & x & x & \checkmark & \checkmark & x & Rich semantic labels, multi-route \\
IndustReal \citep{schoonbeek2024industreal} & \checkmark & x & \checkmark & x & x & x & PSR, ego-only, error modeling \\
Ego-Exo4D \citep{grauman2024ego}   & \textbf{($\sim$6\% ind.)} & \checkmark & \checkmark & \checkmark & \checkmark & x & 1,200h skilled activities \\
\hline
\textbf{OpenMarcie (ours)} & \textbf{(100\%)}\checkmark & \checkmark & \checkmark & \checkmark & \checkmark & \checkmark & Industrial multitasking, 8 modalities \\
\hline
\end{tabular}
}
\vspace{-5mm}
\label{table:ComparativeTable}

\end{table*}
The advancement of smart factories depends on the integration of human intelligence into automated systems, promoting a more efficient, adaptive, and human-centered industrial paradigm.
A fundamental component of this integration is human activity recognition (HAR), which enables systems to understand, support, and optimize human motion within complex industrial workflows.
Typically, video data has served as a rich source of valuable information for HAR. 
As a result, extensive research efforts are dedicated to developing innovative vision-based methods and video-based datasets to support the community \citep{caba2015activitynet,2020mmaction2}, including event cameras \citep{wang2024hardvs}. 
\cref{table:ComparativeTable} presents a comparison with state-of-the-art dataset for HAR in industrial environments. 
Despite recent advances, existing HAR datasets in industrial contexts suffer from three major limitations: (1) a lack of truly multimodal data combining wearable sensors, vision, and audio in a synchronized manner, (2) a reliance on highly constrained, protocol-driven tasks, which do not reflect the open-ended, procedural nature of real-world industrial work, and (3) limited demographic diversity or task complexity. 
In general, most datasets focus on short, isolated actions, failing to capture the extended, multi-step activities typical of human workflows in manufacturing.
Furthermore, human action is inherently multimodal, integrating sensory, cognitive, and motor processes. 
Actions depend on visual, auditory, tactile, while cognitive and emotional states shape movement, speech, and expression \citep{schmidt2001human}.
Variability in motion, context dependence, and environmental interactions add further complexity, making multimodal analysis essential for accurate interpretation \citep{bello2023inmyface,bello2024unimodal}.
To effectively understand and replicate human activity, AI and robotics must process diverse data sources, including video, audio, and motion sensors. 
Wearable and multipositional data sources and synchronized visual information (egocentric and exocentric) have proven to be highly valuable \citep{yoshimura2024openpack,dallel2020inhard,zheng2023ha}. 
This is especially true in industrial environments, where vision-only-based methods may raise privacy concerns and increase the risk of technology leakage \citep{bello2025tsak}.
Accordingly, we present OpenMarcie (see \Cref{fig:Marcie1}).
Our contributions are threefold.
First, we introduce OpenMarcie, a multimodal dataset with eight sensing modalities, including LiDAR depth, multi-view pose estimation, object detection, and precise positioning. 
Second, the dataset captures realistic industrial assembly and disassembly tasks where participant decisions shape alternative routes, enabling the study of procedural knowledge acquisition. 
Third, we provide rich metadata from 36 volunteers, synchronized wearable sensing streams, speech-to-text narrations, and detailed manual annotations that capture overlapping, multi-primitive actions—supporting fine-grained analysis of complex workflows.

\section{OpenMarcie dataset}
\label{sec:openmarcie}
Two primary experimental settings are defined: Ad-hoc and Procedural scenarios, featuring bicycle and 3D printer assembly/disassembly, respectively, as shown in \Cref{fig:Marcie1}. 
The scenarios involved a diverse set of skills, including: Mechanical competencies, such as tool usage, component alignment, and fastening procedures; Cognitive processes, including interpretation of visual instructions or schematics and problem-solving when inconsistencies arise; Fine motor skills, essential for manipulating small or intricate components with precision.
{We select two contrasting tasks: bicycle assembly, a familiar, goal-oriented scenario with open-ended actions, and 3D printer assembly, a highly procedural task requiring interpretation of detailed instructions and unfamiliar components. 
Together, they capture unscripted repair and structured production-line assembly while incorporating broader activities such as wiring, cable routing, pulley installation, unpacking, and part organization. 
This diversity, enables benchmarking across real-world industrial workflows. 
Current benchmarks focus on activity classification, open-vocabulary captioning, and cross-modal alignment, establishing strong baselines for the dataset’s scope and utility.
For both scenarios, 3 ZED X AI stereo cameras without polarizers are distributed as best as possible to cover the entire view of the experiment's room. 
\Cref{fig:Marcie2} depicts the experimental room setting with example views of the exocentric cameras for the scenario \textbf{(a)} ad-hoc bicycle assembly and \textbf{(b)} procedural scenario 3D printer experiment.
ZED stereo cameras come with a rich SDK and allow customized AI methods to be included in the pipeline. 
In \cref{tab:SensingModalities} the recorded sensing modalities for both scenarios are presented. 
In total, the number of raw channels available is around 282.
This includes various sensor placements. 
Furthermore, each participant reviewed and signed an ethical agreement for each scenario, ensuring compliance with the Declaration of Helsinki.
The Ethical Board of the German Research Center for Artificial Intelligence approved the study under agreement numbers HRW-35/24 and SMD-30/24.

\subsection{Ad-hoc Scenario (a): Bicycle}

The assembly and disassembly of a bicycle is proposed as a structured, goal-oriented task that allows participants to autonomously determine their approach. 
It reflects the nature of many practical, outcome-driven scenarios encountered in both industrial and experimental settings.
Common errors, such as improperly tightened bolts or misaligned brakes, are generally straightforward to detect and rectify, contributing to the task’s safety and pedagogical value. 
Due to the widespread familiarity and accessibility of bicycles, this task is especially relevant for use in training environments, educational contexts, and research applications, including studies on human-robot collaboration and task planning. 
The volunteers were equipped with the wearable sensor suite illustrated in \Cref{fig:Marcie3}. 
The setup includes inertial measurement units (IMUs) and barometric pressure sensors mounted on both hands and on the head (integrated into a glasses frame). 
Additionally, thermal and spectrometer sensors are positioned on the chest and shoulder. 
The system is complemented by an egocentric RGB-D camera, which includes LiDAR-based depth sensing.
\Cref{fig:Marcie3} also presents sample visualizations of the multimodal sensor data collected during the task.
For ground truth, we selected the most informative exocentric view (typically the door-side camera in bicycle assembly; \Cref{fig:Marcie2}) and we (humans) manually annotated it using an intent-aware verb–object–tool scheme (\Cref{fig:Marcie4}). This yields semantically meaningful segments aligned with task intent, including multi-label cases (e.g., walking while carrying). Each annotation (Verb, Tools, Object, Remarks) is temporally aligned across all modalities (egocentric video, inertial, thermal, audio). We additionally generate concise natural-language descriptions with GPT-4o \citep{achiam2023gpt}(see \cref{sec:SectionAnnotation}), providing soft labels that improve consistency and interpretability while preserving human-created ground truth.

\subsection{Procedural Scenario (b): 3D Printer}
\begin{figure*}[]
    \centering
\includegraphics[width=0.95\linewidth]{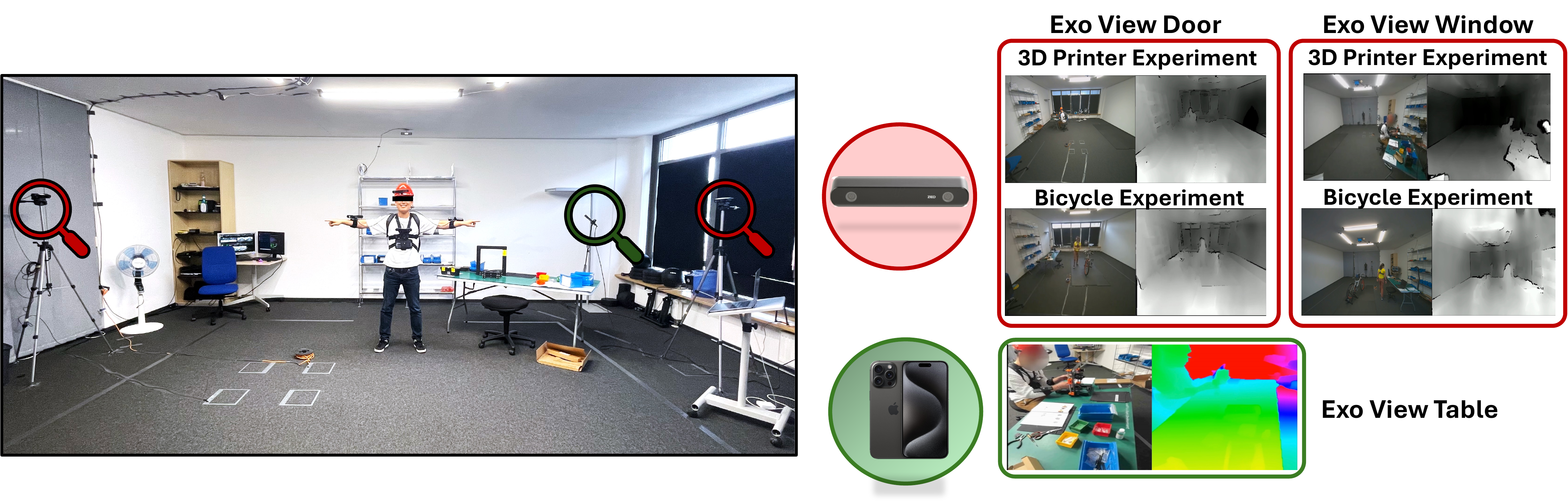}
    \caption{Experiment room setting with example views of the exocentric RGBD cameras.}
    \vspace{-5mm}
    \label{fig:Marcie2}
\end{figure*}

\begin{table}[t]
\centering
\footnotesize
\caption{Sensing modalities information for each scenario}
\setlength{\tabcolsep}{3pt}
\renewcommand{\arraystretch}{0.8}
\begin{tabular}{p{1.9cm} p{2.1cm} p{2.6cm} c}
\hline
Scenario & Modality & Position & Ch \\
\hline
\multirow{8}{*}{Ad-Hoc \textbf{(a)}} 
  & IMU            & Wrists, Forehead            & 21 \\
  & Magnetometer   & Wrists, Forehead            & 9 \\
  & Barometer      & Wrists, Head                & 3 \\
  & Temperature    & Wrists, Forehead            & 3 \\
  & Spectrometer   & Chest, R. Shoulder          & 24 \\
  & Thermal Camera & Chest, R. Shoulder          & (8×8)×2 \\
  & RGB-Lidar      & Chest                        & 4 \\
  & Stereo Sound   & Chest                        & 2 \\
  & RGBD           & 2 Exocentric cameras         & 8 \\
\hline
\multirow{8}{*}{Procedural \textbf{(b)}}
  & IMU            & Wrists                      & 20 \\
  & Magnetometer   & Wrists                      & 6 \\
  & Barometer      & Wrists                      & 2 \\
  & Temperature    & Wrists, Forehead, Chest     & 4 \\
  & RGBD-IMU       & Forehead, Chest             & 28 \\
  & RGB-LiDAR      & Chest, Table                & 8 \\
  & Stereo Sound   & Chest, Table                & 4 \\
  & RGBD           & 2 Exocentric cameras        & 8 \\
\hline
Both Scenarios & 17 modalities & 9 positions & 282 Ch \\
\hline
\end{tabular}
\label{tab:SensingModalities}
\vspace{-10pt}
\end{table}

Building a 3D printer requires significant cognitive effort to interpret written and video instructions. 
Additionally, it involves handling both small components, such as screws, and larger hardware parts like the metal main frame. 
Therefore, this scenario provides an ideal setting for monitoring human action recognition, closely resembling tasks commonly seen in industrial assembly lines.
The Assemble Yourself 3D printer original Prusa i3 MK3S+ kit is used for this scenario. 
The creators offer a detailed set of assembly instructions, with each plastic part accompanied by its 3D model STL file for easy reprinting.
This also means that the 3D models, combined with OpenMarcie’s egocentric videos, can enhance object detection through a predefined dictionary of parts.
This approach intuitively monitors the printer’s assembly status and guides the builder toward the correct next step, accelerating the construction process.
Before the experiment, each volunteer is required to watch a 30-minute video of the complete Prusa assembly guide, available online at \url{https://www.youtube.com/watch?v=uToqSlh64R4}. 
This ensures they understand the key aspects of the task without needing further clarification or instructions from the observer, allowing them to assemble the printer independently.
The idea is for one volunteer to start the assembly using only out-of-the-box instructions and settings, with approximately one hour to build as much as possible. 
The next volunteer would then continue from where the previous participant left off, and this process would repeat sequentially. 
This setup requires each subsequent participant to assess the current assembly status, understand the previous volunteer's progress, and determine how best to proceed, increasing the cognitive challenge to the task.
The participants wore 4 ZED 2 - AI Stereo Cameras. 
Two ZEDs were mounted on the wrists (used only for IMU and barometer signals, with stereo cameras disabled), one on the forehead via a helmet, and one on the chest via a strap to capture visual streams. 
On the person's chest, an iPhone 15 Pro is set up with a cellphone cover with the camera and the Lidar sensor facing the front.
The data is transferred by USB cables to a Jetson Orin AGX, which is placed inside a backpack together with a Green Cell Powerbank 26800mAh 128W to seamlessly record the data for one hour by a volunteer. 
During the experiment, an external human observer provided real-time narration of the participants' actions and contextual details of the scene.
The audio recordings of these narrations were subsequently transcribed using faster-whisper (v1.1.1) and ctranslate2 (v4.4.0), employing the "large-v3" model configuration \citep{radford2022whisper}.
Examples of the soft labels generated using the methods are depicted in \Cref{fig:Marcie4} \textbf{Scenario (b)}. 
The hard labels were then obtained via a two-stage pipeline.
The process involves iteratively identifying and refining discrete action classes using Deepseek-r1 \citep{guo2025deepseek}, then using GPT-4o \citep{achiam2023gpt} with prompt engineering to convert soft-label sentences into persistent hard labels for model training.  
This process—consisting of human annotation followed by a two-stage LLM-based translation—yields our final hard labels for downstream model training.
In Scenario (b), we adopt a two-stage pipeline leveraging transcribed narration, unlike Scenario (a). 
Despite this methodological difference, both yield a consistent multimodal annotation format. 
To validate LLM-generated labels, we perform bidirectional consistency checks (structured→caption→structured), achieving strong alignment (Macro F1 = 0.715 in Scenario a; METEOR = 0.531 in Scenario b; see \cref{sec:SectionAnnotation}). 
These results support the reliability of LLMs as structured translators in our partially supervised pipeline.
Moreover, sensor placements were adapted to suit the specific ergonomics and task demands of each scenario. 
While this design results in some variation—particularly between Scenario (a) (bicycle assembly) and Scenario (b) (3D printer assembly)—key wearable modalities remain consistent across both tasks. 
Most notably, inertial measurement units (IMUs) are placed on the wrists, wearable LiDAR units are mounted on the chest, and stereo microphones are consistently used. 
These shared sensor positions enable meaningful multimodal comparisons, particularly for HAR and captioning benchmarks. 
As illustrated in \ref{fig:Marcie3}, this overlap is visually evident and supports comparative analysis despite scenario-specific adaptations.

\begin{figure*}[]
    \centering
    \includegraphics[width=0.85\linewidth]{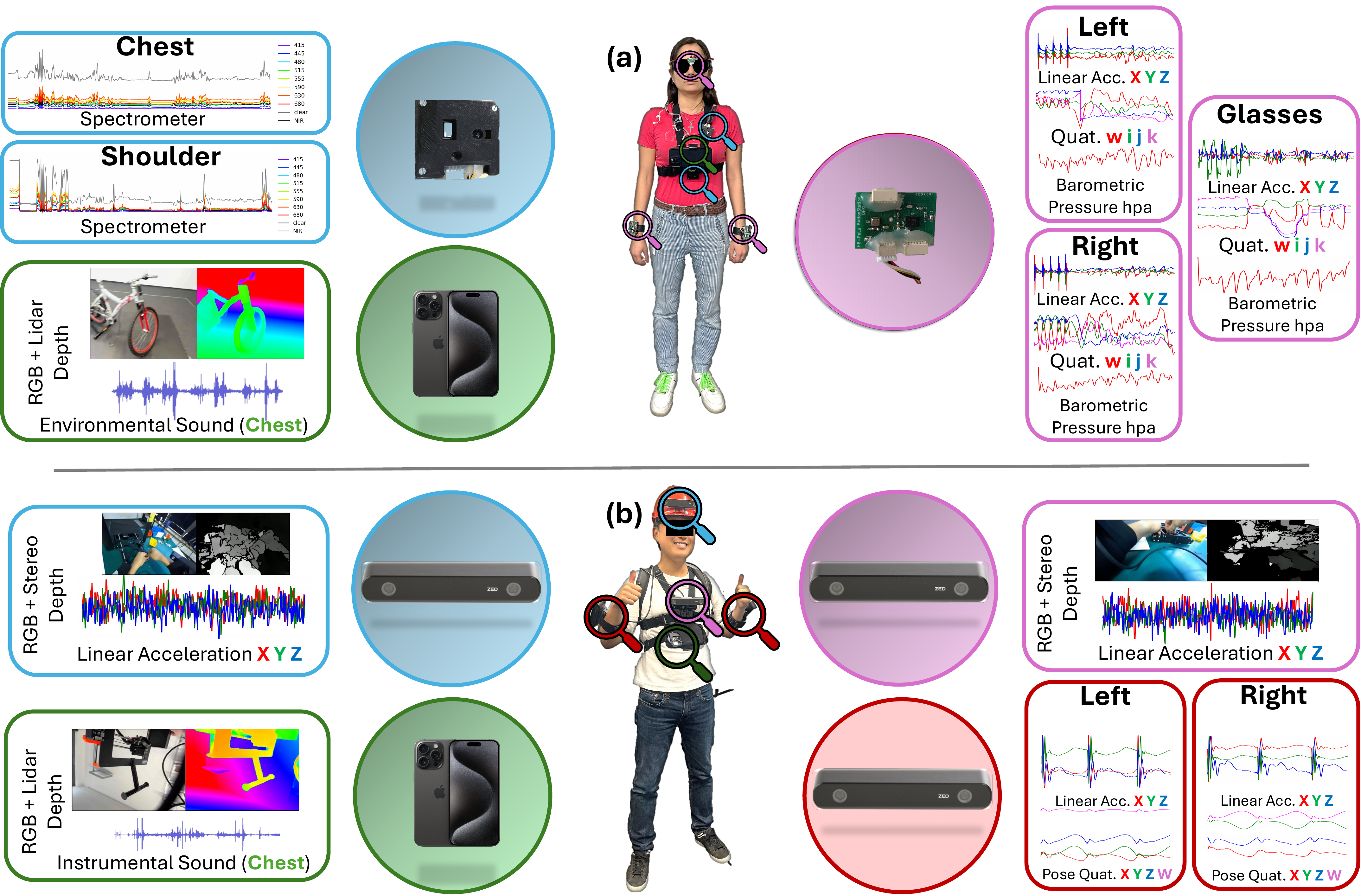}
    \caption{Participant wearable setup and sensor signals. \textbf{(a)} Participant in the ad-hoc Scenario: bicycle assembly. \textbf{(b)} Participant in the procedural Scenario: 3D printer assembly. Wrist-mounted ZEDs provided only IMU and barometer data (stereo cameras disabled), while head and chest cameras captured visual streams. Sensor placements were adapted to the ergonomics and task demands of each scenario.}
    \label{fig:Marcie3}
 
\end{figure*}

\begin{figure*}
    \centering
    \includegraphics[width=0.85\linewidth]{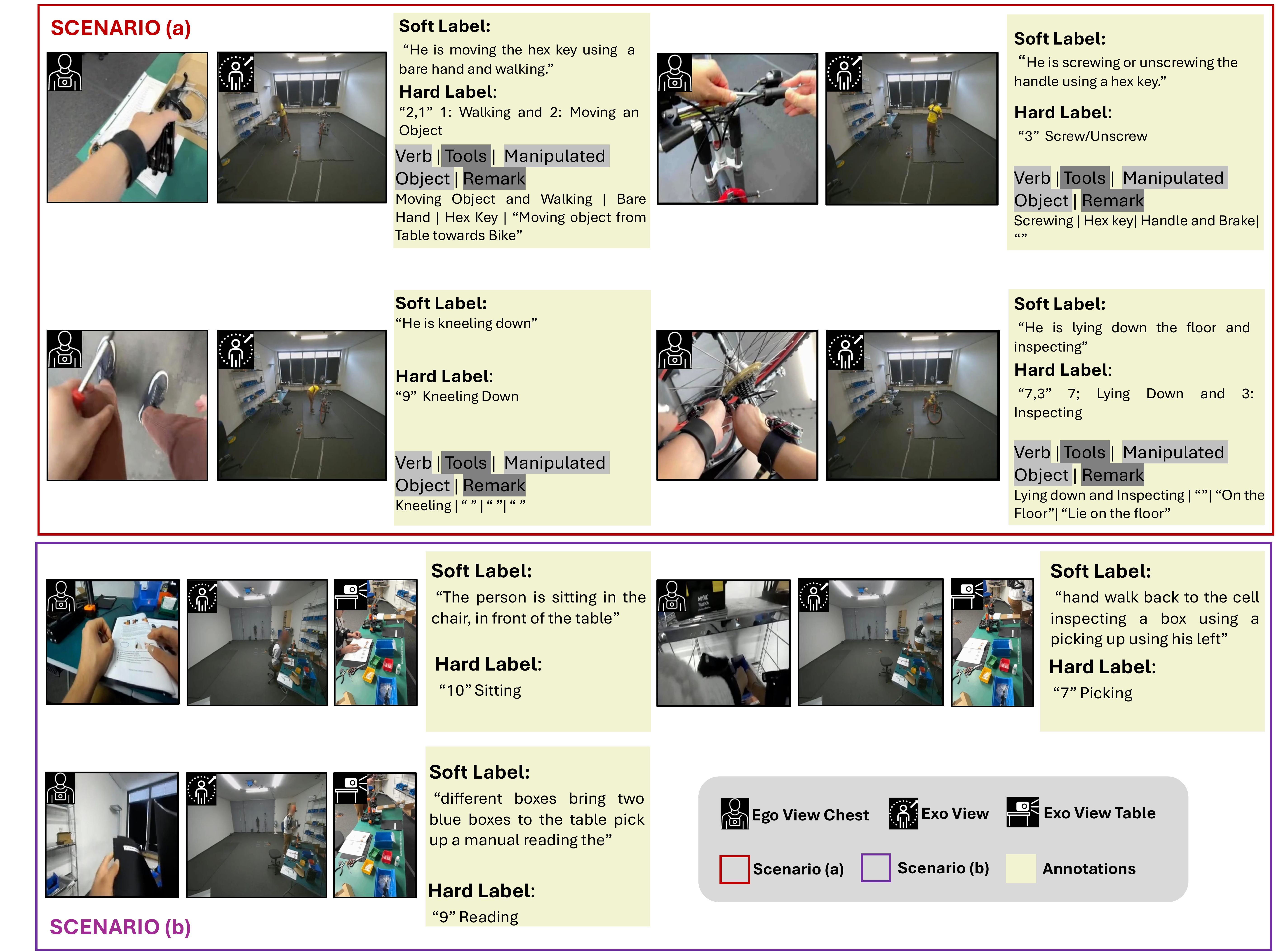}
    \caption{Egocentric and exocentric views of activity examples, accompanied by annotations (see \cref{sec:SectionAnnotation}). 
    These include both soft and hard labels for two scenarios: \textbf{(a)} the ad-hoc bicycle assembly/disassembly task, and \textbf{(b)} the procedural 3D printer construction task.}
    \label{fig:Marcie4}
 
\end{figure*}

\section{Statistic}
\begin{figure*}[]
    \centering
    \includegraphics[width=0.9\linewidth]{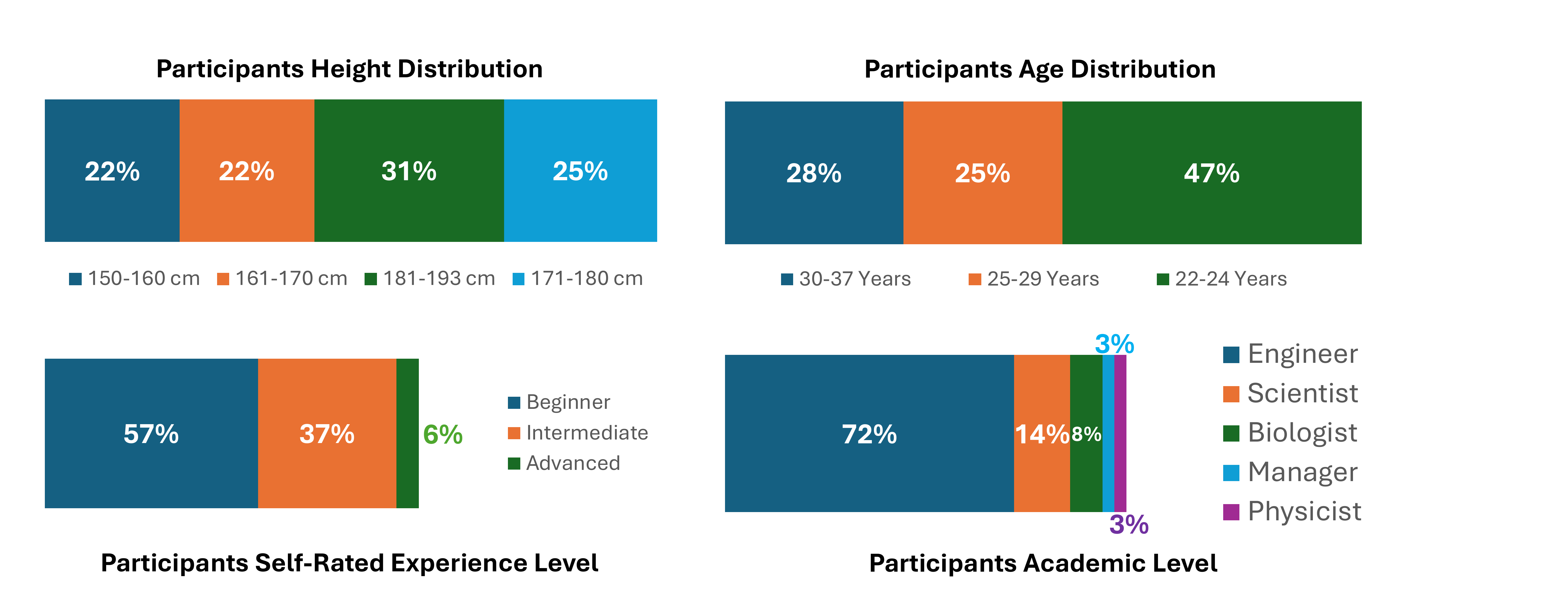}
    \caption{Participants statistic. \textbf{Top left} Height distribution ranging from 150 to 193 cm. \textbf{Top right} Age distribution spanning 22 to 37 years. \textbf{Bottom left} Self-reported experience levels in assembly tasks, categorized as beginner, intermediate, and advanced. \textbf{Bottom right} academic level of participants, with engineers representing the majority and managerial roles accounting for only 3\% of the sample.}
    \label{fig:Marcie5}
    \vspace{-5mm}
\end{figure*}
The study involved a diverse group of participants whose demographic and professional characteristics are summarized in \Cref{fig:Marcie5}. 
In terms of height, participants ranged from 150 to 193 cm, with the largest group (31\%) falling within the 181–193 cm range. The age distribution was skewed toward younger individuals, with 47\% aged 22–24, 25\% aged 25–29, and 28\% aged 30–37.
Participants self-reported their experience levels in assembly-related tasks: 57\% identified as beginners, 37\% as intermediate, and 6\% as advanced. 
Regarding academic and professional background, engineers represented the majority at 72\%, followed by computer scientists (14\%), biologists (8\%), physicists (3\%), and managers (3\%).
This composition reflects a technically oriented, predominantly early-career participant pool with varying levels of practical experience in assembly tasks.
The international diversity of the participants, who come from more than 20 countries, contributes to the cultural and experiential diversity of the study.
Moreover, the majority of participants (31 individuals) were right-handed, while a smaller subset (5 individuals) identified as left-handed. 
While the participant cohort predominantly comprises right-handed individuals with engineering backgrounds—potentially limiting demographic generalization—this demographic reflects the typical industrial workforce, thus enhancing ecological validity.
Participants reported their ownership and usage habits related to bicycles and 3D printers (see \Cref{fig:Marcie7}). 
A majority indicated that they owned a bicycle, while only a small fraction owned a 3D printer. 
Usage patterns reflected this ownership disparity: bicycles were used more frequently than 3D printers, with participants reporting daily, weekly, or monthly use. 
In contrast, the majority of participants reported either very infrequent use or no use of 3D printers.
These differences highlight a greater familiarity and hands-on experience with bicycles among participants, making bicycle-related tasks more intuitive for the group, while 3D printer tasks likely required more instruction or exploration due to limited prior exposure.
The participant characteristics provide valuable context for interpreting task performance and interaction behavior observed during the study. 
The combination of technical backgrounds, varied assembly experience levels, and broad international representation contributes to a rich and diverse dataset. 
Furthermore, the participants’ greater familiarity with bicycles compared to 3D printers offers a natural contrast between ad-hoc and procedural tasks, reinforcing the relevance of the chosen scenarios for studying goal-oriented activities across different levels of prior knowledge and skill.
\Cref{fig:Marcie8} represents an example of activity distribution within OpenMarcie. 
These actions represent a diverse set of activities that humans are expected to perform in industrial scenarios, specifically during assembly tasks.
The dataset focuses on two distinct scenarios. 
\cref{fig:Marcie8} \textbf{Left} is the distribution of activities for the ad-hoc bicycle scenario (\textbf{Scenario (a)}). 
The most frequent activities are screwing/unscrewing and inspecting, as these actions are an integral part of handling bicycle parts. 
In contrast, hammering or lying down occurred at more   defined stages of the assembly process. 
\cref{fig:Marcie8} \textbf{Right} shows the distribution of activities for the 3D printer assembly procedural scenario. 
Moving objects (“Move"), Adjust, and Read are the dominant categories. 
In contrast to the ad-hoc scenario, the Unlabeled class is less prominent in the 3D printer case.

\begin{figure*}[]
    \centering
    \includegraphics[width=0.7\linewidth]{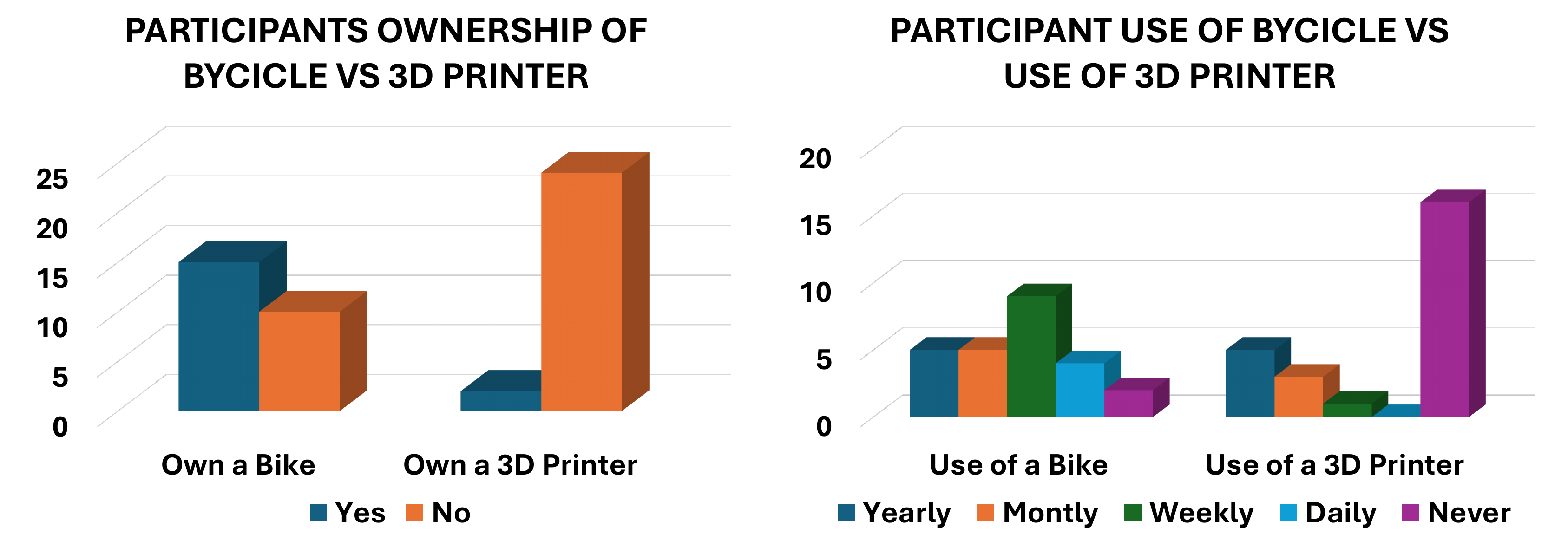}
    \caption{Participants' ownership and usage habits related to bicycles and 3D printers.}
    \label{fig:Marcie7}
    
\end{figure*}

\begin{figure*}[]
    \centering
    \includegraphics[width=\linewidth]{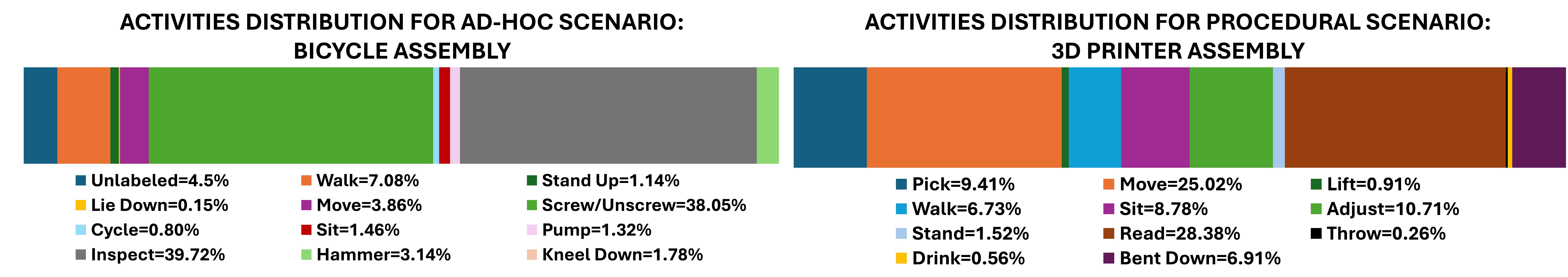}
    \caption{OpenMarcie activities distribution for the ad-hoc scenario and the procedural scenario.}
    \label{fig:Marcie8}
    \vspace{-5mm}
\end{figure*}


\section{Validation Benchmarks}
\begin{table}[]
    \caption{Macro F1 scores for human activity recognition in Scenario (a): Bicycle Assembly and Scenario (b): 3D Printer Assembly, with and without the null class.}

    \footnotesize
    \centering
    \setlength{\tabcolsep}{3pt}
    \renewcommand{\arraystretch}{0.95}
    \begin{tabular}{lcccc}
    \hline
    \multirow{2}{*}{\textbf{Modality}} & \multicolumn{2}{c}{\textbf{Scenario (a)}} & \multicolumn{2}{c}{\textbf{Scenario (b)}} \\
     & \multicolumn{4}{c}{\textbf{Macro F1 ($\uparrow$})} \\
     & No Null &  Null & No Null &  Null\\
    \hline
     Inertial(I) &0.834$\pm$0.007&0.811$\pm$0.007 & 0.750$\pm$0.015&0.674$\pm$0.003\\ 
     Acoustic(A) & 0.489$\pm$0.018& 0.469$\pm$0.017 & 0.425$\pm$0.004&0.432$\pm$0.005 \\
     Vision(V) & 0.757$\pm$0.011 & 0.729$\pm$0.011 &0.705$\pm$0.004 &0.655$\pm$0.003\\
     I + A & 0.803$\pm$0.012&0.782$\pm$0.010 & 0.744$\pm$0.004&0.666$\pm$0.003\\ 
     A + V &0.739$\pm$0.016&0.714$\pm$0.013 &0.695$\pm$0.003 &0.646$\pm$0.003 \\
     I + V & \textbf{0.882}$\pm$\textbf{0.009}& \textbf{0.851}$\pm$\textbf{0.009} & \textbf{0.773}$\pm$\textbf{0.000}&\textbf{0.685}$\pm$\textbf{0.000}\\
     I + A + V& 0.859$\pm$0.010& 0.831$\pm$0.011 & 0.763$\pm$0.003&0.676$\pm$0.003\\
    \hline
    \end{tabular}
    \label{tab:HAR}
 \vspace{-10pt}
\end{table}

\begin{table}[]
    \caption{Open vocabulary captioning results for Scenario (a): Bicycle Assembly and Scenario (b): 3D Printer Assembly, including cosine similarity values with and without the null class.}
    \footnotesize
    \centering
    \setlength{\tabcolsep}{3pt}
    \renewcommand{\arraystretch}{0.95}
    \begin{tabular}{lcccc}
    \hline
    \multirow{2}{*}{\textbf{Modality}} & \multicolumn{2}{c}{\textbf{Scenario (a)}} & \multicolumn{2}{c}{\textbf{Scenario (b)}} \\
     & \multicolumn{4}{c}{\textbf{Cosine Similarity ($\uparrow$})} \\
     & No Null & Null & No Null & Null\\
    \hline
     Inertial(I) &0.518$\pm$0.023&0.501$\pm$0.022 & 0.642$\pm$0.002&0.640$\pm$0.002\\ 
     Acoustic(A) & 0.361$\pm$0.030& 0.341$\pm$0.018 &0.316$\pm$0.003 & 0.323$\pm$0.004\\
     Vision(V) & 0.479$\pm$0.016 & 0.463$\pm$0.014 & 0.632$\pm$0.002&0.631$\pm$0.003\\
     I + A& 0.512$\pm$0.021&0.493$\pm$0.020 & 0.644$\pm$0.002&0.641$\pm$0.003\\ 
     A + V&0.466$\pm$0.025&0.444$\pm$0.012 & 0.626$\pm$0.003& 0.625$\pm$0.004\\
     I + V& \textbf{0.561}$\pm$\textbf{0.016}& \textbf{0.531}$\pm$\textbf{0.014} & \textbf{0.655}$\pm$\textbf{0.000}&\textbf{0.655}$\pm$\textbf{0.000}\\
     I + A + V & 0.547$\pm$0.020& 0.519$\pm$0.017 &0.647$\pm$0.001&0.646$\pm$0.003\\
    \hline
    \end{tabular}
    \vspace{-5mm}
    \label{tab:OpenVoca}
\end{table}
To assess the utility and versatility of OpenMarcie, we establish validation benchmarks across three core tasks: Human Activity Recognition (HAR), Open Vocabulary Captioning, and Cross-Modal Alignment. 
These tasks are chosen to reflect key challenges in industrial and task-driven environments. 
HAR supports downstream applications like safety monitoring, skill assessment, and robotic imitation. 
Open vocabulary captioning enables procedural documentation and naturalistic human-robot interaction. 
Cross-modal alignment is essential for sensor substitution, retrieval, and transfer in environments where sensing modalities may be limited or asynchronous. 
These tasks also address known gaps in prior datasets—such as short action windows, lack of verbal grounding, and limited sensor diversity—positioning OpenMarcie as a testbed for real-world, multimodal intelligence.
\paragraph{Human Activity Recognition}
in industrial settings is challenging due to procedural, goal-driven, and concurrent activities. 
\begin{table*}[t!]
\footnotesize
\centering
\caption{Cross-modal alignment results for Scenario (a): Bicycle Assembly and Scenario (b): 3D Printer Assembly.}
\setlength{\tabcolsep}{3pt}
\renewcommand{\arraystretch}{0.9}
\begin{tabular}{lcccccc}
\hline
\multirow{2}{*}{\textbf{Modality}} &
\multicolumn{3}{c}{\textbf{Scenario (a)}} &
\multicolumn{3}{c}{\textbf{Scenario (b)}} \\
& R@1 & R@5 & Top-1 & R@1 & R@5 & Top-1 \\
\hline
Inertial (I) + Text (T)          & 0.324$\pm$0.016 & 0.655$\pm$0.025 & 0.481$\pm$0.018 & 0.312$\pm$0.016 & 0.642$\pm$0.026 & 0.468$\pm$0.019 \\
Acoustic (A) + T                 & 0.241$\pm$0.014 & 0.583$\pm$0.025 & 0.342$\pm$0.016 & 0.227$\pm$0.013 & 0.567$\pm$0.022 & 0.329$\pm$0.015 \\
Vision (V) + T                   & 0.437$\pm$0.015 & 0.768$\pm$0.017 & 0.556$\pm$0.016 & 0.421$\pm$0.013 & 0.751$\pm$0.018 & 0.541$\pm$0.014 \\
I + A + T                        & 0.347$\pm$0.014 & 0.679$\pm$0.019 & 0.495$\pm$0.017 & 0.334$\pm$0.015 & 0.663$\pm$0.017 & 0.479$\pm$0.018 \\
A + V + T                        & 0.412$\pm$0.013 & 0.740$\pm$0.020 & 0.533$\pm$0.015 & 0.395$\pm$0.014 & 0.723$\pm$0.019 & 0.517$\pm$0.014 \\
I + V + T                        & \textbf{0.485}$\pm$\textbf{0.014} & \textbf{0.803}$\pm$\textbf{0.019} & \textbf{0.587}$\pm$\textbf{0.016} 
                                 & \textbf{0.467}$\pm$\textbf{0.013} & \textbf{0.787}$\pm$\textbf{0.015} & \textbf{0.570}$\pm$\textbf{0.016} \\
I + A + V + T                    & 0.470$\pm$0.015 & 0.795$\pm$0.019 & 0.579$\pm$0.016 & 0.453$\pm$0.014 & 0.779$\pm$0.018 & 0.563$\pm$0.016 \\
\hline
\end{tabular}
\vspace{-5mm}
\label{tab:Cross}
\end{table*}

Unlike prior datasets \citep{yoshimura2024openpack} with short, scripted actions, OpenMarcie captures long, unsegmented sequences with natural variation, tool use, and diverse action paths. 
Its rich multimodal setup with multipositional sensors, synchronized ego/exo views, and verbal narrations offers a realistic testbed for human-centered automation.
We evaluate HAR using three modalities: egocentric video, right-hand IMU, and instrumental sound.
Each is tested independently with a modality-specific model, ViT \citep{dosovitskiy2020image} for video, DeepConvLSTM \citep{singh2020deep} for IMU, and EnCodec \citep{defossez2022high} with a temporal classifier for audio on a 12-class activity task with subject-disjoint splits.
For multimodal evaluation, we explore all pairwise combinations and full three-way fusion using a late-fusion transformer \citep{pandeya2021deep} that integrates temporally encoded embeddings from each stream. 
This setup quantifies both the individual discriminative power and the complementarity across modalities.
\cref{tab:HAR} shows Macro F1 scores across modalities for two scenarios (a, b) with and without null class inclusion. 
Inertial + Vision consistently outperforms all other combinations, achieving the highest Macro F1 scores. 
Multimodal fusion generally improves performance over unimodal inputs, especially over Acoustic alone, which performs the worst across conditions.
\paragraph{Open Vocabulary Captioning}
generates free-form descriptions of actions and scenes, crucial for industrial tasks like documentation, operator feedback, and human-machine handovers.
Unlike prior datasets focused on short, generic activities with templated labels \citep{liu2020manifesting}, OpenMarcie provides unscripted spoken narrations aligned with visual and audio streams. 
This enables grounded language generation in fine-grained, goal-oriented settings and allows investigation of how multimodal context, especially verbal intent, enhances captioning performance.
Similar to HAR, we use modality-specific encoders to regress sentence embeddings derived from participant narrations as proposed in OV-HAR \citep{ray2025initial}. 
Captioning is framed as embedding prediction, with outputs decoded via Vec2Text \citep{morris2023text} using embedding retrieval. 
We evaluate unimodal, pairwise, and fused setups. 
This approach enables efficient, open-vocabulary captioning without large language models.
\cref{tab:OpenVoca} reports cosine similarity scores for captioning, showing trends similar to HAR. 
Inertial + Vision achieves the best performance across both datasets, while Acoustic alone performs the worst. 
Adding Acoustic to other modalities provides marginal or no improvement, indicating its limited contribution compared to Inertial and Vision.
\paragraph{Cross Modal Alignment}
aims to learn a shared representation across sensory modalities, enabling retrieval, recognition, and transfer. 
While models like CLIP \citep{radford2021learning}, Multi³Net \citep{fortes2024enhancing} succeed on web-scale data, they lack structured, sensor-rich recordings. 
OpenMarcie offers a realistic testbed with synchronized video, audio, and IMU from goal-driven tasks, supporting alignment studies in complex, real-world settings beyond internet-scale benchmarks.
Inspired by ImageBind \citep{girdhar2023imagebind}, we use contrastive learning to align egocentric video, IMU, audio, and language into a shared embedding space. 
Each modality is encoded with a dedicated backbone and trained with a multi-modal InfoNCE loss over synchronized positive pairs. 
We evaluate all 2, 3, and 4 modality combinations, enabling fine-grained analysis of alignment and shared information across modalities.
As given in \cref{tab:Cross}, for both scenarios, combining inertial and vision modalities with text yields the strongest alignment performance. 
Vision and text perform well alone; while acoustic features contribute less individually, their performance improves when combined with other modalities.
Overall, richer modality fusion improves alignment, though gains taper with full combinations.
Embedding-based approaches enable efficient, scalable evaluation across tasks, highlighting OpenMarcie’s value for studying real-world, sensor-rich human activity understanding.
Across all tasks, HAR, open vocabulary captioning, and cross-modal alignment, multimodal fusion consistently outperforms unimodal inputs, with inertial and visual modalities providing the strongest signal. 
Audio alone provides limited information but contributes meaningfully when combined with other modalities.
Moreover, because data was collected in a test-bench rather than a real factory, it lacks authentic industrial noise (e.g., machinery, vibrations). 
While constrained, the modality remains useful in multimodal fusion and highlights opportunities for privacy-preserving sensing under realistic conditions. 
Additional experiments further probing the causes of audio’s limited performance are provided in the supplementary material.

\section{Conclusion}

OpenMarcie is a large-scale multimodal dataset for human activity recognition, cross-modal learning, and industrial automation. 
It captures both ad-hoc and procedural assembly tasks with synchronized egocentric/exocentric video, wearable sensors, anonymized audio, and textual narration, enabling robust benchmarking for embodied AI. 
The dataset reflects realistic industrial workflows through sequential, collaborative activities involving error correction and task monitoring. 
While participants are predominantly right-handed engineers—limiting demographic generalization—this aligns with typical industrial workforces, enhancing ecological validity. Annotations cover only part of the dataset’s potential, as its multi-view recordings support future labeling of objects, actions, interactions, and poses. 
Sensor placements were ergonomically adapted per scenario but key modalities (e.g., wrist IMUs, chest-mounted LiDAR, stereo microphones) remain consistent, allowing multimodal comparisons. 
Overall, it provides a flexible foundation for research in fine-grained activity recognition, human–robot collaboration, and context-aware AI in industrial settings.

\section*{Acknowledgments}
This work was supported by the German Federal Ministry of Education and Research (BMBF) through the Cross-Act project (Grant Agreement No. 01IW25001).
This work was also supported by the IITP grant funded by the MSIT (RS-2019-II190079, AI Graduate School Program (Korea University) and IITP-2026-RS-2024-00436857, IITP-ITRC), and by the Technology Innovation Program (RS-2025-25453819) funded by the MOTIR.  

\FloatBarrier
\newpage
{
    \small
    \bibliographystyle{ieeenat_fullname}
    \bibliography{references}
}

%
%

\clearpage
\setcounter{page}{1}
\maketitlesupplementary

This supplementary document contains additional information about OpenMarcie.
\cref{ComparisonDataset} provides a detailed comparison to prior datasets, outlining differences in sensing coverage, industrial fidelity, and annotation design. 
\Cref{sec:SectionUser} presents the metadata information about each participant independently. 
\Cref{sec:Ethical} declares the ethical consideration and societal impact of the OpenMarcie dataset with emphasis in user privacy and anonymization. 
\Cref{sec:SectionAnnotation} describes how a large language model semantically transforms human-written descriptions and vice-versa, serving as a structured translator rather than a primary labeling agent.
\Cref{sec:object} shows the object distribution for both experimental scenarios and illustrates example scenes with object segmentation and human skeleton poses.
\Cref{sec:tool_handling} reports tool-handling statistics in the bicycle scenario and highlights the multimodal cues involved in tool-contact events. 
\Cref{sec:SectionBench} details the implementation of the proposed benchmarks and presents the evaluation metrics used.
\cref{AudioAdditional} reports additional audio-only experiments designed to probe the modality’s limitations under different conditions, and analyzes audio’s role in tool-contact detection. 
Specifically, we (1) evaluate the full pipeline with pre-anonymized (raw) audio streams to isolate the effect of the anonymization process, and (2) replace Encodec embeddings with a non–ML-based mel-spectrogram classifier to disentangle the influence of the embedding method from the content itself.
\Cref{sec:SensorRationale} describes the rationale behind the selected sensor modalities and their complementary physical signals.
\cref{sec:modern_architectures} reports results from modern multimodal architectures, including Perceiver IO and AnyMAL, demonstrating improvements over late-fusion baselines.
\cref{FuturereResearch} outlines several research directions that extend beyond the benchmarks presented in this work.

\section{Detailed Comparison to Prior Datasets, Scope, and Generalization}
\label{ComparisonDataset}
To provide a clearer perspective, we include a narrative comparison of each cited dataset, highlighting how their design choices converge with or diverge from OpenMarcie in terms of industrial fidelity, sensing breadth, and suitability for advanced research tasks.
\begin{itemize}
    \item InHARD \cite{dallel2020inhard}, developed for human–robot collaboration, integrates three exocentric RGB views with wearable inertial mocap, making it one of the few industrial datasets to include wearable sensing. 
    Its scope, however, is limited to single-action recognition in fixed exocentric settings. 
    In contrast, OpenMarcie extends beyond this by pairing wearables with synchronized egocentric and exocentric video and by providing multi-action labels, thereby capturing overlapping activities that more faithfully reflect real-world industrial workflows.
    
    \item LARa \cite{niemann2020lara} captures logistics activities using IMUs, optical mocap, and a single RGB camera, providing valuable insight into worker variability in warehouse settings. 
    Its scope, however, is primarily exocentric and lacks the multimodal depth of OpenMarcie. 
    In the absence of egocentric streams or concurrent multi-action labels, LARa is best suited for controlled HAR scenarios.
    OpenMarcie advances this space by integrating egocentric video, wearable sensing, and multi-route task design, thereby enabling richer supervision for collaborative and multitasking contexts.
    
    \item OpenPack \cite{yoshimura2024openpack} is a large-scale logistics dataset comprising 53.8 hours of recordings that integrate IMUs, 2D keypoints, depth, and IoT data. 
    Its primary strength lies in combining wearable sensing with process-level logging, though its perspective remains exclusively exocentric. 
    OpenMarcie addresses these gaps by providing egocentric multimodal video aligned with exocentric multiviews and by introducing explicit multi-action annotation. 
    This design enables models not only to classify activities but also to reason about simultaneity and task intent.
    
    \item Assembly101 \cite{sener2022assembly101} offers multi-view egocentric and exocentric video with dense procedural annotations, establishing a strong benchmark for vision-based activity understanding. 
    However, it does not incorporate wearable sensing or overlapping activity labels. 
    OpenMarcie extends this direction by retaining ego–exo coverage while adding wearable IMUs, audio, thermal, and spectrometer signals, and by explicitly modeling concurrent actions. 
    These additions make OpenMarcie particularly well-suited for advancing cross-modal learning.
    
    \item IKEA-ASM \cite{ben2021ikea} is a furniture assembly dataset that provides RGB-D multiview recordings with annotations for atomic actions and manipulated objects. 
    Its main strength is the inclusion of depth and pose data from multiple exocentric Kinect sensors, but it lacks egocentric perspectives and wearable sensing. 
    OpenMarcie complements this resource by combining egocentric video with wearables and additional non-visual sensors, thereby supporting cross-modal transfer beyond RGB-D alone.
    
    \item HA4M \cite{cicirelli2022ha4m} focuses on gear-train assembly and provides six modalities captured with Azure Kinect, including RGB, depth, IR, and skeleton data.
    While it emphasizes vision-rich multimodality, it is restricted to exocentric viewpoints and does not include wearable sensing. 
    OpenMarcie advances this space by integrating egocentric and exocentric cameras with wearables and multi-action labels, thereby supporting more realistic multitasking scenarios in industrial workflows.
    
    \item HA-ViD \cite{zheng2023ha} provides fine-grained multi-view assembly recordings with detailed annotations covering subjects, verbs, objects, tools, and collaboration cues. 
    It offers rich semantic supervision and supports alternative task routes.
    However, it does not include wearable sensing or egocentric perspectives. 
    OpenMarcie complements HA-ViD by incorporating egocentric video, wearable streams, and explicit concurrent multi-action labels, thereby extending applicability to multitasking and multimodal alignment.
    
    \item IndustReal \cite{schoonbeek2024industreal} is designed for Procedure Step Recognition (PSR), using egocentric video to capture procedural errors and flexible subgoals. 
    Its distinctive contribution is its focus on error modeling, though it remains limited to ego-only recordings without wearable sensing or multiview exocentric support. 
    OpenMarcie, while currently benchmarked on HAR, captioning, and cross-modal alignment, integrates egocentric and exocentric video with wearables and multi-action labels, providing a complementary foundation for PSR and error modeling—both of which are included in our staged roadmap.
    
    \item Ego-Exo4D \cite{grauman2024ego} is a large-scale dataset comprising over 1,200 hours of skilled activities, captured with synchronized egocentric and exocentric video, audio, IMU, gaze, and language streams.
    It is unmatched in scale and modality breadth, yet only a small portion of the recordings are industrial. 
    OpenMarcie takes a complementary approach by focusing exclusively on industrial workflows, augmenting them with multi-action annotations and specialized wearables. 
    In this way, OpenMarcie serves as a domain-focused counterpart to Ego-Exo4D, prioritizing depth over scale in factory settings.
\end{itemize}

OpenMarcie is the only dataset to jointly provide wearables, egocentric and exocentric multiview recordings, explicit concurrent multi-action labels, and complete industrial coverage (see \cref{table:ComparativeTable}). 
While prior datasets emphasize individual strengths—such as rich semantic annotations (HA-ViD), large scale (Ego-Exo4D), wearable integration (InHARD, LARa, OpenPack), or multi-view video (Assembly101, IKEA-ASM)—none combine all of these elements within an industrial setting. 
This unique positioning makes OpenMarcie particularly well-suited for the advanced benchmarks outlined in our roadmap, including procedural planning, skill assessment, intent prediction, fine-grained segmentation, pose reasoning, cross-modal transfer, and cross-modal generation.

\subsection{Scope and Industrial Positioning}

OpenMarcie focuses specifically on assembly-centric industrial workflows rather than attempting exhaustive coverage of all industrial activities. 
The dataset is structured around two complementary regimes: (i) ad-hoc, experience-driven assembly and disassembly (bicycle), capturing exploration, corrective behavior, and goal-oriented problem solving; and (ii) instruction-following procedural assembly (3D printer), reflecting standardized production and onboarding scenarios.

While we do not claim coverage of activities such as large-scale packaging, heavy machinery inspection, or conveyor-based logistics, the procedural scenario naturally incorporates industrially relevant behaviors beyond assembly itself, including unpacking, part organization, manual reading, workspace preparation, object transport, and sequential collaborative continuation of partially completed tasks. 
These behaviors reflect structured production-line dynamics within a controlled environment.

The design choice prioritizes controlled variability and multimodal richness over maximal task breadth, enabling systematic study of modality complementarity and procedural reasoning within assembly workflows.

\subsection{Ecological Validity and Controlled Test-Bench Environment}

Data collection was conducted in a controlled test-bench setting rather than an operational factory. 
Consequently, factors such as heavy machinery noise, large-scale environmental clutter, strict safety constraints, and unplanned workflow interruptions are underrepresented. 
We acknowledge this as a limitation.

However, the controlled environment provides several methodological advantages: synchronized multimodal capture, reduced confounding variables, stable camera calibration, and high annotation fidelity. 
This setup enables precise temporal alignment across modalities and systematic investigation of cross-modal robustness, which would be significantly more difficult in uncontrolled factory conditions.

OpenMarcie should therefore be understood as a controlled yet industrially inspired benchmark for multimodal perception in assembly contexts, rather than as a direct replica of a live factory floor.

\subsection{Generalization and Domain Shift Considerations}

With respect to generalization, models trained on OpenMarcie are expected to transfer most directly to assembly-oriented tasks involving fine motor manipulation, procedural reasoning, and multimodal perception under moderate environmental variability. 
Transfer to substantially different industrial domains—such as high-noise machining environments, high-speed packaging lines, or large-scale quality inspection pipelines—is not guaranteed.

Future evaluations should explicitly investigate robustness under domain shift conditions, including simulated machinery noise injection, environmental clutter augmentation, and cross-site validation in operational factory environments. 
Such experiments would provide quantitative assessment of real-world transferability.

In this context, OpenMarcie serves as a structured foundation for studying multimodal robustness, modality complementarity, and procedural activity modeling in assembly-centric industrial workflows, while recognizing the boundaries of its ecological validity.

\section{User Metadata}
\label{sec:SectionUser}

Based on \Cref{fig:Marcie6}, \Cref{tab:VolunteerBike}, and \Cref{tab:Volunteer3D}, the participant data across the two scenarios—bicycle assembly (Scenario (a)) and 3D printer assembly (Scenario (b))—reflects a diverse and well-balanced cohort in terms of demographics and professional backgrounds. 
Scenario (a) includes 12 participants, while Scenario (b) features 24, offering a broader basis for analysis.
Males form the majority in both groups, although Scenario (b) exhibits greater gender diversity, with a higher number of female participants.

The majority of participants are right-handed, with only four identifying as left-handed, as shown in the summarized visualization, indicating a strong dominance of right-handed individuals across the dataset. 
Participants range in age from their early 20s to late 30s, and most hold academic degrees in engineering. 
Additional disciplines represented include computer science, biology, physics, and management, demonstrating multidisciplinary relevance.

Geographically, the dataset includes participants from multiple continents—South America, Asia, Europe, Africa, and North America—highlighting broad international representation. 
Scenario (b) shows particularly rich demographic variety, with over 15 distinct national origins. 
Experience levels, self-reported on a scale from 1 to 3, vary across participants, with most indicating beginner to intermediate familiarity with the task domain.

Overall, the metadata illustrates the inclusiveness and diversity of the participant pool, supporting the dataset’s utility for developing generalizable models in embodied AI, human-robot interaction, and vision-based behavior analysis.

In addition to the demographic information, we also report upper-body anthropometric measurements for Scenario (b), summarized in Table~\ref{tab:UpperBodyAnthropometry}.

These measurements include shoulder-to-shoulder (SS) and shoulder-to-wrist lengths (SR-WR and SL-WL), expressed in centimeters, and provide a quantitative characterization of participant body proportions.

The physical dimensions of a person in a video—particularly their absolute scale, bounding box size, and relative segment proportions—are highly relevant to pose estimation performance. 
Although many modern algorithms aim to be scale-invariant, the pixel-level representation of the body directly influences keypoint localization accuracy and subsequent 3D reconstruction. 
In our setup, cameras are installed in the environment to estimate pose while participants perform assembly tasks and move freely within the room, continuously changing their distance to the cameras and thus their apparent scale. 
Providing subject-specific anthropometric measurements therefore enables more precise metric reconstruction in a centimeter-based reference frame, supports accurate multi-camera triangulation, and offers reliable ground-truth information for evaluating and improving pose estimation methods.

\begin{figure}[h]
    \centering
    \includegraphics[width=0.8\linewidth]{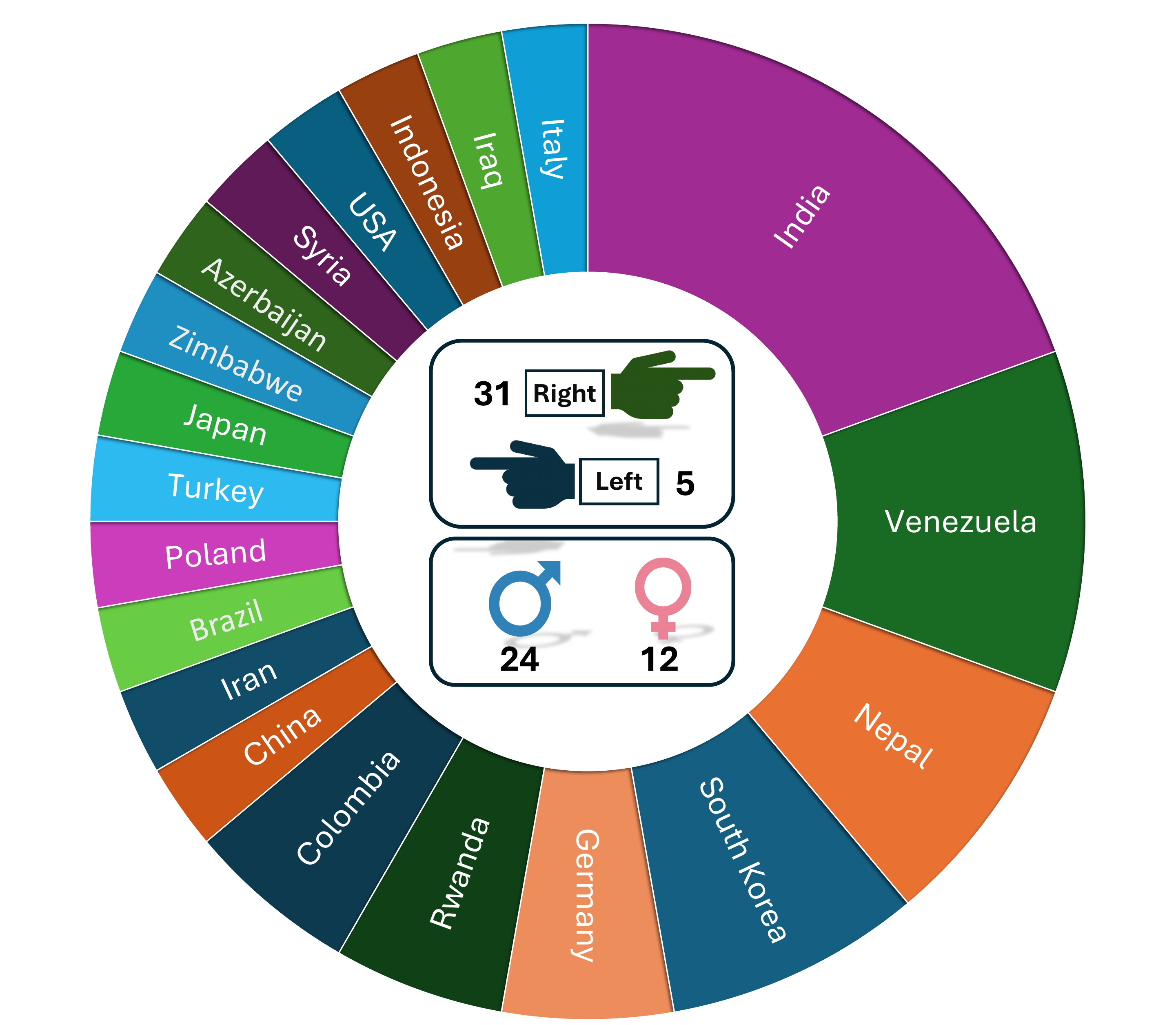}
    \caption{Distribution of participants by nationality, dominant hand and self identified sex.}
    \label{fig:Marcie6}
\end{figure}

\begin{table*}[h]
    \centering
    \caption{Participants' metadata information for the Ad-hoc Scenario (a).}
    \resizebox{\textwidth}{!}{
    \begin{tabular}{c c c c c c c c}
        \hline
         ID& Sex &Age &Height & Dominant Hand &Academic Level &Demographic & Experience Level (1-3)\\
         \hline
         P1-B& F& 33 & 160 & R& Engineer & South America (Venezuela)& 2\\
         P2-B& F& 26 & 176 & R& Engineer & Europe (Poland)& 3\\
         P3-B& M& 29 &175  & R& Computer Scientist & Europe (Germany)& 1\\
         P4-B& M& 26 & 175 & L& Engineer & South Asia (India)& 1\\
         P5-B& M& 33 & 189 & R& Computer Scientist & South America (Brazil)& 1\\
         P6-B& M& 36 & 188 & R& Engineer & East Africa (Rwanda)& 1\\
         P7-B& M& 25 & 168 & R& Engineer & South Asia (India)& 1\\
         P8-B& M& 37 & 178 & R& Engineer &East Asia (South Korea)& 2\\
         P9-B& M& 27 & 176  & R& Engineer & Middle East (Iran)&2 \\
         P10-B&M& 27 & 174  & R& Engineer & South Asia(India)&1 \\
         P11-B& M&34  & 193  & R& Engineer & South America (Venezuela)& 1\\
         P12-B& M&30  & 160  & R& Engineer & South East Asia (China)& 1\\       
         \hline
    \end{tabular}
    }
    \label{tab:VolunteerBike}
\end{table*}

\begin{table*}[t!]
    \centering
    \caption{Participants metadata information for the procedural Scenario (b).}
    \resizebox{\textwidth}{!}{
    \begin{tabular}{c c c c c c c c}
        \hline
         ID& Gender &Age &Height & Dominant Hand &Academic Level &Demographic & Experience Level (1-3)\\
         \hline
         P1-D& M &37  &178  &R &Engineer  & East Asia (South Korea)& 2\\
         P2-D& F & 25 &159  & R& Engineer  & South East Asia Pacific (Nepal)&2 \\
         P3-D& M & 34 & 193 & R& Engineer & South America (Venezuela)& 1 \\
         P4-D& F & 26 &165  &R &Engineer  &Europe (Italy) &1 \\
         P5-D& F& 33 &160  &R & Engineer &South America (Venezuela) & 2\\
         P6-D& F & 25 &164  &R & Engineer & South Asia (India)& 1\\
         P7-D& M& 24  &175  & R &Engineer  & Middle East (Iraq)& 3 \\
         P8-D& M & 25  &168  &R &Engineer  & South Asia (India)&1 \\
         P9-D& F& 25 & 162 & R& Physicist &South America (Colombia) &2 \\
         P10-D& F &  27&  155&R &Engineer  &South East Asia (Indonesia) & 2\\
         P11-D& F & 24 &160  &R &Engineer  & South Asia (India) &1 \\
         P12-D& M &36  &188  &R &Engineer  &East Africa (Rwanda) &1 \\
         P13-D& M &24  &168  &R &Biologist  &North America (USA) & 2 \\
         P14-D& M&22  &187  &R &Management  & Middle East (Syria)&2 \\
         P15-D& F& 24 &150  &R & Engineer  & Caucasus Asia (Azerbaijan) &1 \\
         P16-D&M & 25 &165  &R &Engineer  &South Africa (Zimbabwe) &1 \\
         P17-D& M& 22 &167  &R &Engineer  &South East Asia Pacific (Nepal) &2 \\
         P18-D& M& 29  &175  &L & Computer Scientist &Europe (Germany) &1 \\
         P19-D& F &23  &161  &R &Biologist  &East Asia (Japan) &1 \\
         P20-D& M&26  &175  &L & Engineer  &South Asia (India) &1 \\
         P21-D& M& 27 &182  &R & Computer Scientist & East Asia (South Korea) &2\\
         P22-D& M &24  &183  &L &Engineer  & South East Asia Pacific (Nepal)& 2\\
         P23-D& F& 24 &156  &R & Microbiologist &South America (Colombia) & 1\\
         P24-D& M& 25 & 178 & L&  Computer Scientist& Middle East (Turkey) & 1\\
         \hline
      
    \end{tabular}
    }
    \label{tab:Volunteer3D}
\end{table*}

\begin{table}[t!]
    \centering
    \caption{Upper-body anthropometric measurements: Shoulder-to-Shoulder length (SS), Shoulder-Right to Wrist-Right length (SR-WR), and Shoulder-Left to Wrist-Left length (SL-WL), measured in cm. Values are presented for each participant along with Mean ± Standard Deviation.}
    \resizebox{0.4\textwidth}{!}{
    \begin{tabular}{c c c c}
        \hline
        ID & SS (cm) & SR-WR (cm) & SL-WL (cm) \\
        \hline
        P1-D & 51 & 55 & 55 \\
        P2-D & 34 & 49 & 48 \\
        P3-D & 51 & 61 & 61 \\
        P4-D & 36 & 57 & 57 \\
        P5-D & 36 & 54 & 54 \\
        P6-D & 38 & 53 & 53 \\
        P7-D & 46 & 59 & 59 \\
        P8-D & 44 & 55 & 54 \\
        P9-D & 40 & 55 & 54 \\
        P10-D & 36 & 50 & 50 \\
        P11-D & 38 & 53 & 53 \\
        P12-D & 47 & 65 & 64 \\
        P13-D & 44 & 55 & 55 \\
        P14-D & 50 & 61 & 61 \\
        P15-D & 38 & 46 & 46 \\
        P16-D & 45 & 54 & 54 \\
        P17-D & 41 & 56 & 57 \\
        P18-D & 43 & 56 & 56 \\
        P19-D & 32 & 52 & 52 \\
        P20-D & 42 & 56 & 56 \\
        P21-D & 47 & 59 & 59 \\
        P22-D & 47 & 57 & 58 \\
        P23-D & 38 & 49 & 49 \\
        P24-D & 42 & 57 & 57 \\
        \hline
        Mean ± SD & 41.33 ± 5.82 & 55.04 ± 4.42 & 54.96 ± 4.43 \\
        \hline
    \end{tabular}
    }
    \label{tab:UpperBodyAnthropometry}
\end{table}

\begin{figure*}[t!]
    \centering
    \includegraphics[width=0.8\linewidth]{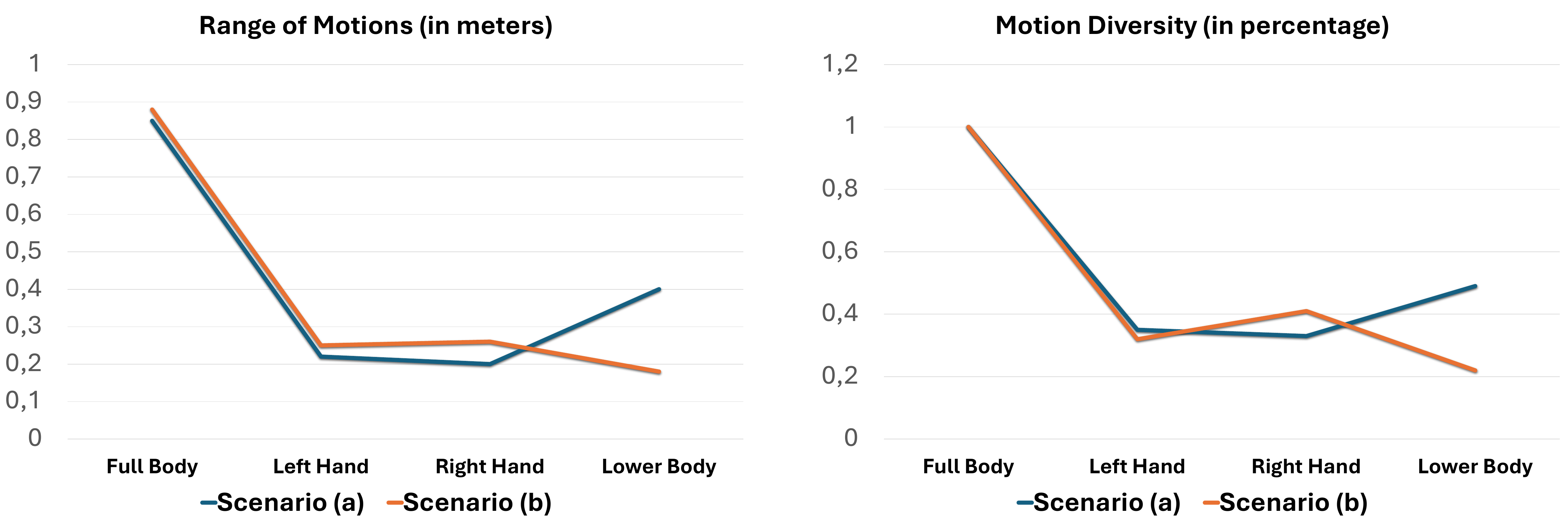}
    \caption{Comparison of range and diversity of body motions across Scenario (a): bicycle assembly and Scenario (b): 3D printer assembly.}
    \label{fig:Marcie9}
\end{figure*}

\Cref{fig:Marcie9} presents a comparative analysis between range of motion (Left) and motion diversity (Right) across two distinct assembly scenarios: Scenario (a), Bicycle assembly, and Scenario (b), 3D Printer assembly. 
Each metric is broken down by body region: Full Body, Left Hand, Right Hand, and Lower Body. 
Scenario (b) exhibits a greater range of motion compared to Scenario (a), particularly for the full body as well as the left and right hands. 
This increased movement may be attributed to the multiple instances of searching around the table, shelf, and floor observed in Scenario (b).
Scenario (a) involves kneeling, lying on the ground, and standing up to inspect specific areas of the bicycle. Consequently, it demonstrates a greater lower body range of motion and movement diversity compared to the 3D printer scenario (Scenario b).

Table \ref{tab:activitiesbike} presents a list of acronyms used to represent various activities and objects involved in Scenario (a): Bicycle assembly. 
The activities are denoted by single-letter acronyms such as W for Walking, M for Move (manipulating an object before the next action), U for Screw/Unscrew, and C for Cycling. 
Other physical actions include S for Sitting down, P for Pumping air into the tires, I for Inspecting with hands or eyes, H for Hammering, K for Kneeling down, A for Standing up, L for Lying down, and T for Cutting. 
Additionally, object-related acronyms are listed, including hx for Hex key, wr for Wrench, sd for Screwdriver, hm for Hammer, sc for Scissors, pl for Plier, pu for Pump, and bh for Bare Hands. 
This table helps clarify shorthand notations used to describe the detailed steps and tools involved in the bicycle assembly process.
\begin{table*}[h]
\footnotesize
\caption{Activities and objects acronyms for the Scenario (a): Bicycle assembly.}
    \centering
    \begin{tabular}{c c }
          \hline
        Acronym & Meaning  \\
        \hline
        W & Walking   \\
        M & Move: Manipulating an object until the next
        action \\
        U & Screw/Unscrew\\
        C & Cycling\\
        S & Sitting down\\
        P&  Pump: Pumping air into the bicycle tires\\
        I & Inspect: Inspecting an object with hand/eyes\\
        H & Hammering \\
        K & Kneeling down \\
        A & Standing up \\
        L & Lie Down \\
        T & Cutting \\
        hx & Hex key\\
        wr & Wrench \\
        sd & Screwdriver\\
        hm & Hammer\\
        sc & Scissors \\ 
        pl & Plier \\
        pu & Pump\\
        bh & Bare Hands \\
          \hline
        
    \end{tabular}
    \label{tab:activitiesbike}
\end{table*}

\section{Ethical Consideration and Societal Impact}
\label{sec:Ethical}
The OpenMarcie dataset was developed with a strong emphasis on ethical research practices and societal responsibility. 
All participants provided informed consent in compliance with the Declaration of Helsinki, and the study was reviewed and approved by the Ethics Board of the German Research Center for Artificial Intelligence under protocols HRW-35/24 and SMD-30/24. 
Participants were informed about the nature of data being collected—including egocentric video, wearable sensor data, and external audio narration, and retained the right to withdraw at any time.
Participants received a 15-euro Amazon voucher for voluntary participation in Scenario (b). 
No compensation was provided for participation in Scenario (a).

To mitigate privacy risks, egocentric recordings were deliberately framed to capture task execution while minimizing exposure of participant identities. 
Verbal narrations were provided by an external observer to avoid capturing participants’ private speech.
These narrations were then transcribed using  faster-whisper (v1.1.1) and ctranslate2 (v4.4.0), employing the "large-v3" model
configuration \cite{radford2022whisper}, and only the resulting text descriptions were used in the benchmark tasks and released in OpenMarcie.

Facial features and biometric identifiers were excluded from all labeling and analysis processes.
A two-stage anonymization procedure was applied to the video data. 
First, participant faces were automatically detected and blurred using the open-source deface Python package (threshold set at 0.7) \cite{CenterFace,orbhd_deface}.
Second, all videos were manually reviewed, and any remaining visible facial features were fully obscured using DaVinci Resolve \cite{davinci_resolve}.
For audio, only the instrumental components were released. 
Voices were intentionally removed from the audio tracks using the OpenVINO Music Separation plugin in Audacity \cite{intel_openvino_audacity}, which separates recordings into vocal and instrumental stems.
OpenMarcie includes only the instrumental tracks.

OpenMarcie is designed not only to advance research in human activity recognition but also to support transparency and fairness in machine learning. 
The dataset includes comprehensive metadata on participants' demographics, academic background, dominant hand, and self-assessed skill levels, enabling researchers to perform subgroup analyses and audit for potential biases. 
This supports the development of equitable multimodal systems in industrial and embodied AI applications.

From a broader societal perspective, OpenMarcie has the potential to benefit applications such as human-robot collaboration, workplace safety, ergonomic assessment, and adaptive training systems. 
Its real-world, goal-driven scenarios are ideal for advancing models that interpret human actions in complex environments. 
However, such systems may also carry risks if misused—for example, for excessive surveillance or performance monitoring without consent.

We encourage researchers using OpenMarcie to explicitly assess fairness, document performance across diverse subgroups, and consider the downstream implications of deploying human action recognition systems in human-centered settings. 
OpenMarcie aims to support the responsible development of AI by providing a rich yet ethically grounded testbed for real-world multimodal learning.

\section{LLM-based Annotation Translation and Ground Truth Strategy}
\label{sec:SectionAnnotation}
\subsection{LLM Generated Label Validation}
Our pipeline employs GPT-4o to translate human-authored soft activity descriptions into standardized formats—either discrete activity classes (hard labels) or continuous representations (soft labels). 
Importantly, the LLM is not used to generate annotations directly from visual input, but rather to semantically convert existing human-written annotations. 
Thus, it acts as a structured translator rather than a primary labeling agent.

Crucially, this translation process is not fully automated. 
Human oversight is incorporated throughout, particularly during the mapping of soft-label sentences to discrete classes. 
Annotators refine LLM prompts, inspect outputs for ambiguous cases, and resolve semantic inconsistencies. 
This human-in-the-loop approach helps ensure the accuracy and consistency of final labels used for model training.

To validate the quality of LLM-generated annotations, we perform a bidirectional consistency analysis under the two scenarios:

\begin{itemize}
  \item \textbf{Ad hoc scenario (Hard labels $\rightarrow$ Captions $\rightarrow$ Hard labels):} Human-annotated discrete classes are transformed by the LLM into natural-language captions, then back-translated by the LLM into discrete labels.
  \item \textbf{Procedural scenario (Captions $\rightarrow$ Hard labels $\rightarrow$ Captions):} Human written soft label sentences are transformed into discrete classes by the LLM, then regenerated into captions.
\end{itemize}

We measure the consistency between original and recovered labels/captions using Macro F1 Score and METEOR\cite{banerjee2005meteor}.
In both settings, we observe strong alignment between the classification and regression outputs having \textbf{0.715 Macro F1 score} for Scenario (a) and \textbf{0.531 METEOR score} for Scenario (b) respectively, suggesting that LLM-generated labels preserve semantic consistency and structural fidelity. 
This indirect validation supports the utility of LLMs as reliable semantic intermediaries within a partially supervised annotation pipeline.

\subsection{Ad-hoc Scenario (a): Bicycle disassembly/assembly}
For example, given:

\begin{verbatim}
Verb: Moving Object and Walking
Tools: Bare Hand
Manipulated Object: Hex Key
Remarks: "Moving object from 
Table towards Bike"
\end{verbatim}

we issue the following prompt:

\begin{verbatim}
% System message to define assistant
role and style
System: 
You are an expert 
activity‐description assistant.
Always produce a single 
declarative sentence 
in present continuous tense,
third person, starting with 
a capital letter and ending 
with a period.

% User message with the annotation 
payload and example
User:
Convert the following structured 
annotation into 
one clear sentence.

Annotation:
– Verb: Moving Object and Walking
– Tools: Bare Hand
– Manipulated Object: Hex Key
– Remarks: Moving object from Table 
towards Bike

Now, please convert the above annotation.
\end{verbatim}

\texttt{GPT-4o} then output:

\begin{verbatim}
“He is moving the hex key using 
a bare hand and walking.”
\end{verbatim}

This generated sentence is used as the soft label target for downstream model training.

\subsection{Procedural Scenario (b): 3D Printer Assembly/Disassembly}
In the first stage, we used reasoning model DeepSeeker1 \cite{guo2025deepseek} on all soft-label sentences to predict candidate activity classes. 
We iteratively extracted the set of unique predicted classes, re-running the pipeline until convergence (i.e., no new classes appeared). 
Next, we manually verified and merged semantically similar classes to produce the final discrete set: Pick, Move, Lift, Walk, Sit, Adjust, Stand, Read, Throw, Drink, Bent Down.

For example, given the soft label:

\begin{verbatim}
“The person bends down 
to pick something 
off the floor.”
\end{verbatim}

we construct a prompt to DeepSeeker1 that encourages open-ended reasoning about the described action, such as:

\begin{verbatim}
System:
You are a reasoning assistant 
tasked with extracting 
activity-related
verbs or action phrases 
from natural language
descriptions of human behavior.
User:
Extract the key activity-related 
labels 
from the following sentence:
“The person bends down to 
pick something off the floor.”
\end{verbatim}

\noindent
DeepSeeker1 may then return:

\begin{verbatim}
["Pick", "Bend", "Grab"]
\end{verbatim}

Each returned label is treated as a candidate activity class. We check each one against the current working set of known classes. If a label is not already in the set, we add it. This iterative procedure continues over the entire dataset of soft-label sentences. After each pass, we re-run DeepSeeker1 on any newly discovered or ambiguous phrases to catch any missed classes. The process continues until convergence, meaning no new unique classes are added in a full iteration.

In our example, if "Pick" and "Grab" are already in the working set but "Bend" is not, we would update the set as:

\begin{verbatim}
Existing class set: ["Pick", "Grab"
, ...]
Updated class set: ["Pick", "Grab", 
"Bend", ...]
\end{verbatim}

After convergence, we manually verify and merge semantically similar or redundant labels (e.g., merging "Bend" and "BentDown") to finalize a clean, discrete set of activity classes:

\begin{verbatim}
Final class set: ["Pick", "Move", 
"Lift", "Walk", "Sit",
"Adjust", "Stand", "Read", "Throw",
"Drink", "Bent Down"]
\end{verbatim}

In the second stage, we used GPT-4o \cite{achiam2023gpt} with prompt engineering to convert each soft-label sentence into one of these classes, employing a “sticky” logic so that the predicted class persists until a new class is detected at a later timestamp.  

For example, given the soft label: \textit{“The person is sitting in the chair, in front of the table.”}

we issue the following prompt:

\begin{verbatim}
System:
You are an activity-classification 
assistant.
Candidate classes: Pick, Move, Lift, Walk, 
Sit, Adjust, 
Stand, Read, Throw, Drink, BentDown, 
Others.
Sticky logic: retain previous label 
unless a new one is predicted.
User:
Classify the following sentence
into one of the candidate classes:
“The person is sitting in the chair, 
in front of the table.”
\end{verbatim}

\noindent
\texttt{GPT-4o} then outputs: “10 (Sit)”

where the integer “10” refers to the class \texttt{Sit}.
where the integer “10” refers to the class Sit. This two-stage approach yields our final hard labels
for downstream model training.

\paragraph{Ground Truth Acquisition Strategy.}
The ground-truth annotations described above rely on synchronized multi-view capture rather than dedicated motion-capture systems. 
While high-accuracy optical mocap could in principle provide precise kinematic measurements, we intentionally avoided marker-based setups. 
Such systems are intrusive for long-horizon tool use, may restrict natural manipulation, and are particularly susceptible to occlusions during close-range assembly involving frequent hand–object interactions. 
These limitations can compromise ecological validity in assembly-centric workflows. 
Instead, we leverage synchronized multi-view RGB-D acquisition using ZED stereo cameras, which provides strong 3D scene perception while remaining minimally intrusive and scalable. 
Manual, intent-aware annotation is performed on the most informative exocentric view and temporally aligned across modalities, ensuring semantically consistent ground truth while preserving natural interaction behavior.

\paragraph{VLM-Based Automatic Labelling}
\label{sec:vlm_labels}

To assess whether Vision--Language Models (VLMs) can replace or reduce manual annotation effort, we generated automatic soft labels by prompting a state-of-the-art VLM Qwen2-VL \cite{wang2024qwen2} to caption each egocentric video segment using the same temporal boundaries as the manual annotations.
Table~\ref{tab:vlm_openvocab} reports the open-vocabulary cosine similarity achieved when training the regression pipeline on these VLM-generated labels instead of human-written soft labels.

\begin{table}[t]
    \caption{Open-vocabulary cosine similarity when training on VLM auto-labels vs.\ human annotations.  VLM labels yield scores only marginally above chance, substantially below the manual-label baselines.}
    \footnotesize
    \centering
    \setlength{\tabcolsep}{3pt}
    \renewcommand{\arraystretch}{0.95}
    \begin{tabular}{lcccc}
    \hline
    \multirow{2}{*}{\textbf{Modality}} & \multicolumn{2}{c}{\textbf{Scenario (a)}} & \multicolumn{2}{c}{\textbf{Scenario (b)}} \\
     & \multicolumn{4}{c}{\textbf{Cosine Similarity ($\uparrow$})} \\
     & No Null & Null & No Null & Null\\
    \hline
     I  & 0.137$\pm$0.041 & 0.129$\pm$0.038 & 0.168$\pm$0.012 & 0.164$\pm$0.011\\
     A  & 0.098$\pm$0.035 & 0.091$\pm$0.033 & 0.103$\pm$0.009 & 0.099$\pm$0.010\\
     V    & 0.152$\pm$0.029 & 0.143$\pm$0.027 & 0.181$\pm$0.008 & 0.177$\pm$0.009\\
     I + A        & 0.141$\pm$0.039 & 0.133$\pm$0.036 & 0.172$\pm$0.010 & 0.168$\pm$0.011\\
     A + V        & 0.148$\pm$0.033 & 0.139$\pm$0.030 & 0.179$\pm$0.009 & 0.174$\pm$0.009\\
     I + V        & 0.163$\pm$0.028 & 0.154$\pm$0.026 & 0.194$\pm$0.007 & 0.189$\pm$0.008\\
     I + A + V    & 0.159$\pm$0.031 & 0.150$\pm$0.029 & 0.190$\pm$0.008 & 0.185$\pm$0.009\\
    \hline
    \end{tabular}
    \label{tab:vlm_openvocab}
\end{table}

Across all modality combinations and both scenarios, VLM auto-labels produce cosine similarity scores in the range 0.09--0.19, only marginally above a random-embedding baseline ($\approx$0.0).
By comparison, the same regression model trained on human-written soft labels achieves 0.36--0.56 in Scenario~(a) and 0.32--0.66 in Scenario~(b), a gap of 3--4$\times$.
Qualitative inspection reveals that VLM captions tend to describe generic scene context (\textit{``a person standing in a workshop''}) rather than the fine-grained tool--object interactions captured by human annotators (\textit{``He is screwing the brake using a hex key''}).
This discrepancy is especially pronounced for short action segments where the VLM lacks sufficient temporal context to distinguish between visually similar activities.
Given these findings, we omit VLM auto-labels from the main benchmarks.

\section{Object Tracking}
\label{sec:object}
\begin{figure*}[h]
    \centering
    \includegraphics[width=0.8\linewidth]{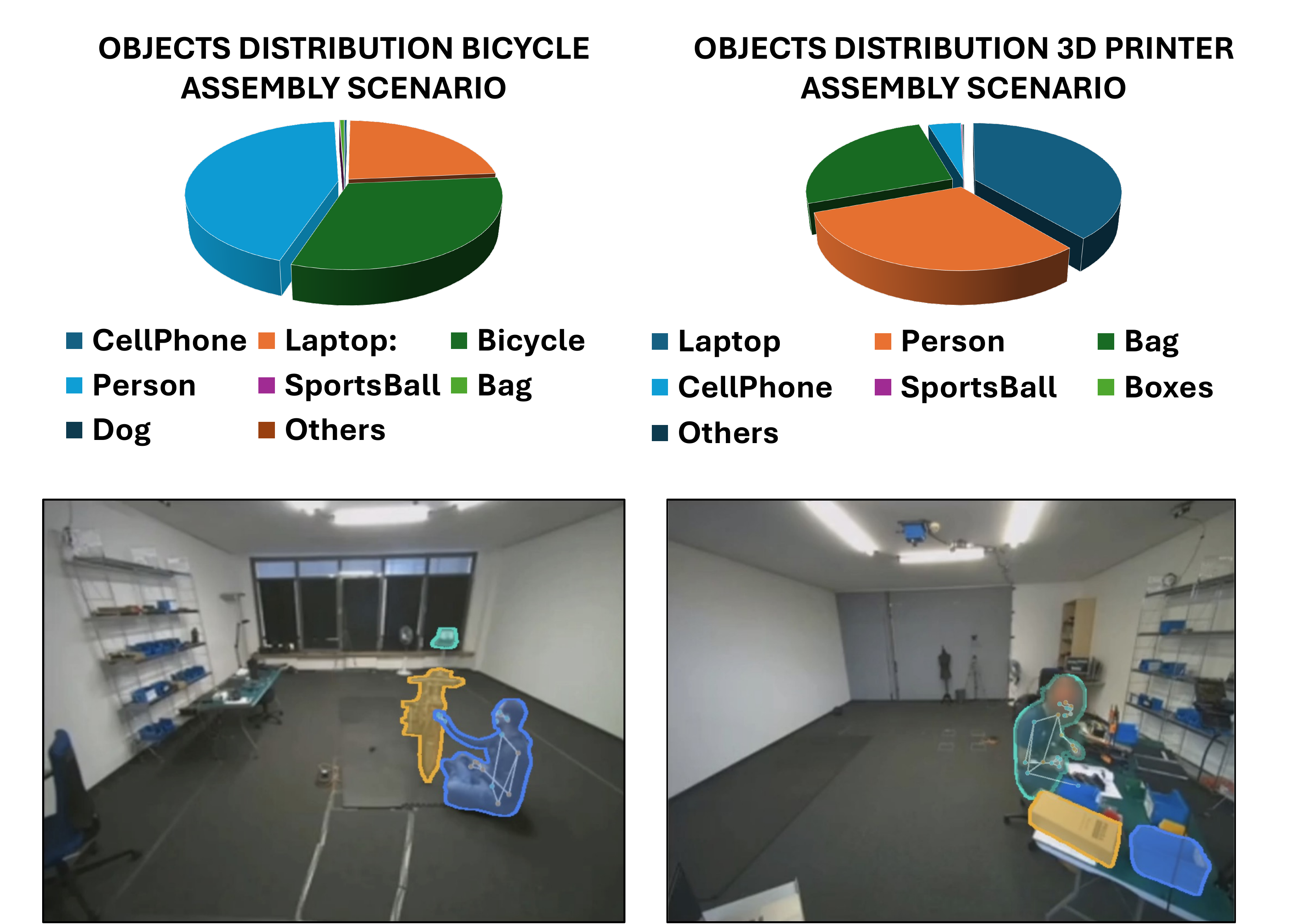}
    \caption{Object distributions in the bicycle and 3D printer scenarios from exocentric views, accompanied by scene visualizations illustrating object segmentation (masks) and human pose estimation in both environments.}
    \label{fig:Marcie10}
\end{figure*}

OpenMarcie provides object segmentation and tracking data from multiple viewpoints, including an exocentric camera and two egocentric positions (head- and chest-mounted wearable cameras) for the 3D assembly experiment (Scenario (b)).
\Cref{fig:Marcie10} \textbf{Top} shows the distribution of detected objects in the Scenario (a): bicycle, and Scenario (b): 3D printer assembly from exocentric views. 
In both cases, "Person" is the most frequently observed category, reflecting the consistent presence of participants during task execution. 
Common objects such as laptops, cellphones, and bags appear in both settings, likely reflecting typical work-related accessories.

Scenario-specific items highlight contextual differences: for example, the bicycle is unique to the bicycle assembly scenario, while boxes—potentially representing packaging or components—are exclusive to the 3D printer setup. 
Less relevant or incidental objects like sports balls and dogs are detected infrequently.
\Cref{fig:Marcie10} \textbf{Bottom} presents example frames with object masks and human skeleton poses for both scenarios.

Overall, the figure underscores context-dependent variations in object presence, providing insights valuable for computer vision and robotics applications.

\section{Tool Handling}
\label{sec:tool_handling}
\begin{table}[t]
    \caption{Tool vocabulary for Scenario~(a) (Bicycle Assembly).  Percentages are computed over the 1{,}769 tool-annotated action instances aggregated across all users.}
    \footnotesize
    \centering
    \setlength{\tabcolsep}{4pt}
    \renewcommand{\arraystretch}{0.95}
    \begin{tabular}{clr}
    \hline
    \textbf{Code} & \textbf{Tool} & Usage in \textbf{\%} \\
    \hline
    bh  & Bare hand                   & 72.5\% \\
    hx  & Hex key         & 16.2\% \\
    sd  & Screwdriver                 & 3.3\%  \\
    wr  & Wrench                      & 2.9\%  \\
    hk  & Hook  & 2.2\%  \\
    pu  & Pump                        & 1.0\%  \\
    pl  & Pliers                      & 1.0\%  \\
    hm  & Hammer                      & 0.6\%  \\
    sc  & Scissors                    & 0.2\%  \\
    \hline
    \end{tabular}
    \label{tab:tool_vocab}
\end{table}

Table~\ref{tab:tool_vocab} lists the nine-tool vocabulary used in the Scenario~(a) verb--tool--object annotation scheme.
Bare-hand manipulation dominates (72.5\%), reflecting the prevalence of positioning and inspection steps during bicycle assembly.
The hex key (Allen key) accounts for 16.2\% of tool instances and is the most frequent specialised tool, consistent with its ubiquitous use in tightening bicycle bolts.
The remaining tools---screwdriver, wrench, hook lever, pump, pliers, hammer, and scissors---collectively cover 11.3\% of instances.
As shown in Table~\ref{tab:tool_contact}, the acoustic modality is especially valuable for detecting these tool-contact events: adding audio to V\,+\,I raises Macro~F1 from 0.905 to 0.923 in Scenario~(a) and from 0.898 to 0.914 in Scenario~(b), because transient impact and friction sounds produced by tools such as the hammer, wrench, and hex key carry discriminative signatures that complement the inertial and visual channels.

\section{Benchmark Architecture}
\label{sec:SectionBench}

\begin{figure*}
    \centering
    \includegraphics[width=0.8\linewidth]{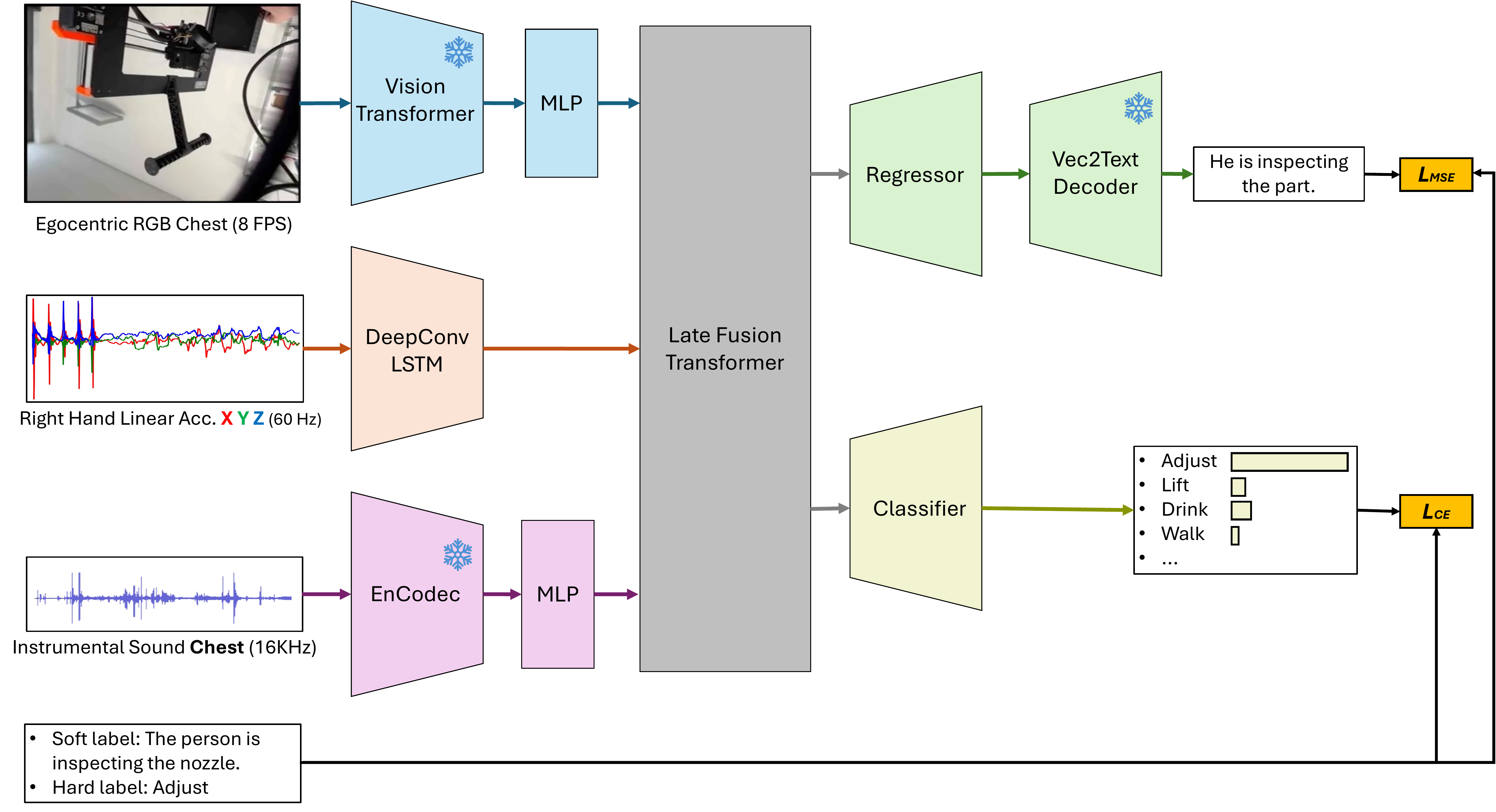}
    \caption{Architectures for classification and regression in human activity recognition and open-vocabulary captioning benchmarks.}
    \label{fig:Marcie11}
\end{figure*}

\begin{figure*}
    \centering
    \includegraphics[width=0.8\linewidth]{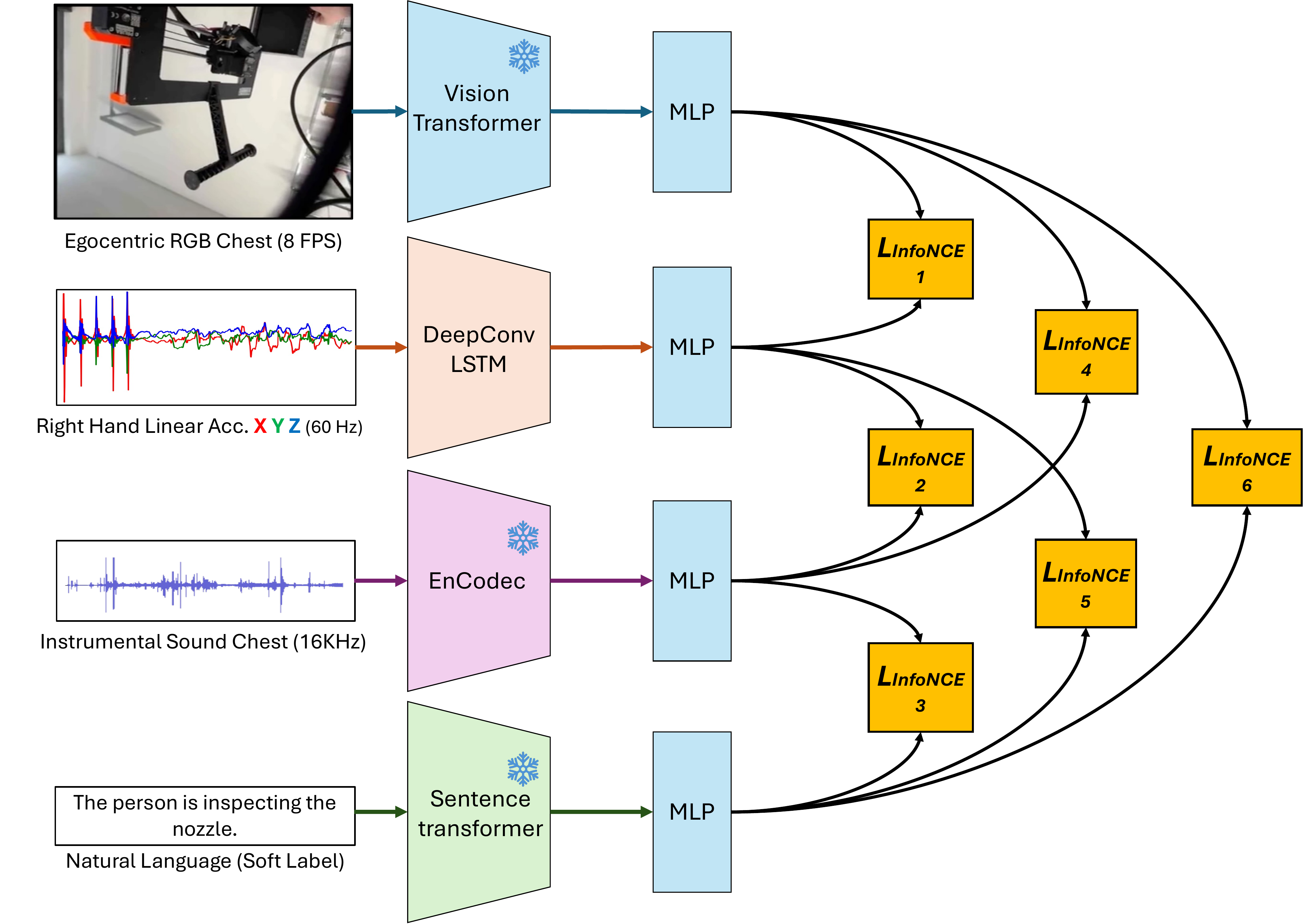}
    \caption{Architecture for cross-modal alignment benchmark.}
    \label{fig:Marcie12}
\end{figure*}

As shown in \Cref{fig:Marcie11}, we segment data into synchronized 1-second windows and evaluate different modality combinations for human activity recognition. 
For each modality, input data is encoded independently: 3 video frames $\{x_v^t\}_{t=1}^3$ are processed via a Vision Transformer \cite{dosovitskiy2020image} $\mathcal{E}_v$, IMU signals $x_i \in \mathbb{R}^{100 \times 6}$ via a DeepConvLSTM \cite{singh2020deep} encoder $\mathcal{E}_i$, and 1-second audio $x_a \in \mathbb{R}^{16000}$ via EnCodec \cite{defossez2022high} $\mathcal{E}_a$, producing embeddings $z_v = \mathcal{E}_v(\{x_v^t\})$, $z_i = \mathcal{E}_i(x_i)$, and $z_a = \mathcal{E}_a(x_a)$, respectively. 
For unimodal models, a classification head $\mathcal{C}$ maps each $z_m$ to logits $\hat{y} = \mathcal{C}(z_m)$, where $m \in \{v, i, a\}$. 
For multimodal combinations (e.g., video+IMU, video+audio, or all three), embeddings are fused via a late-fusion transformer \cite{pandeya2021deep} $\mathcal{F}$: $z = \mathcal{F}(z_{m_1}, z_{m_2}, ...)$, followed by $\hat{y} = \mathcal{C}(z)$. 
All models are trained using cross-entropy loss:
\begin{equation}
    \mathcal{L}_{\text{CE}} = -\sum_{c=1}^{C} y_c \log \hat{y}_c
\end{equation}
where $C = 12$ is the number of activity classes and $y$ is the one-hot ground truth label.

We formulate open vocabulary captioning as a sentence embedding regression task, where models predict language representations from multimodal sensory inputs. 
Each modality is processed independently: 3 sampled video frames $\{x_v^t\}_{t=1}^3$ are passed through a Vision Transformer $\mathcal{E}_v$, 1-second audio $x_a \in \mathbb{R}^{16000}$ through EnCodec $\mathcal{E}_a$, and IMU signals $x_i \in \mathbb{R}^{100 \times 6}$ through DeepConvLSTM $\mathcal{E}_i$, producing embeddings $z_v = \mathcal{E}_v(\{x_v^t\})$, $z_a = \mathcal{E}_a(x_a)$, and $z_i = \mathcal{E}_i(x_i)$. 
For unimodal or multimodal combinations, embeddings are fused via a transformer $\mathcal{F}$ to yield a shared representation $z = \mathcal{F}(z_{m_1}, z_{m_2}, \dots)$. A regression head $\mathcal{R}$ maps $z$ to a predicted sentence embedding $\hat{s} = \mathcal{R}(z)$. 
Ground truth sentence embeddings $s$ are obtained from a pretrained language encoder. 
Models are trained to minimize mean squared error (MSE):
\begin{equation}
    \mathcal{L}_{\text{MSE}} = \| \hat{s} - s \|_2^2
\end{equation}
we perform caption retrieval using a Vec2Text \cite{morris2023text} decoder. 
This embedding-inversion approach enables scalable, low-latency caption generation without autoregressive decoding.

For cross-modal alignment (see \Cref{fig:Marcie12}), we adopt a self-supervised contrastive learning approach using InfoNCE loss to map embeddings from different modalities into a shared representation space similar to ImageBind \cite{girdhar2023imagebind}. 
Using the same modality-specific encoders as before, we compute embeddings $z_m = \mathcal{E}_m(x_m)$ for each modality $m \in \{\text{video, audio, IMU, text}\}$. 
For each temporally aligned pair $(z_m, z_{m'})$, we apply a projection head and compute similarity with other samples in the batch. The InfoNCE loss is given by:
\begin{equation}
    \mathcal{L}_{\text{InfoNCE}} = -\log \frac{\exp(\text{sim}(z_m, z_{m'}) / \tau)}{\sum\limits_{z^-} \exp(\text{sim}(z_m, z^-) / \tau)}
\end{equation}
where $\text{sim}()$ is cosine similarity, $\tau$ is a temperature parameter, and $z^-$ are negative samples.

\subsection{Benchmark Metrics}

We use task-specific metrics suited to the structure and goals of each benchmark:

\textbf{HAR (Macro F1 Score):}  
To account for class imbalance in multi-class activity recognition, we report macro-averaged F1 across all classes:
\begin{equation}
\text{F1}_{\text{macro}} = \frac{1}{C} \sum_{c=1}^{C} \frac{2 \cdot \text{Precision}_c \cdot \text{Recall}_c}{\text{Precision}_c + \text{Recall}_c}
\end{equation}
This gives equal weight to each class regardless of frequency.

\textbf{Captioning (Cosine Similarity):}  
We evaluate caption quality via cosine similarity between predicted and ground truth sentence embeddings:
\begin{equation}
\text{sim}(\hat{s}, s) = \frac{\hat{s} \cdot s}{\|\hat{s}\| \cdot \|s\|}
\end{equation}
This reflects how well the model captures semantic similarity in the shared embedding space, without relying on exact lexical matches.

\textbf{Cross-Modal Alignment (Retrieval Metrics):}  
We assess alignment by retrieving the correct paired modality using Recall@k and Top-1 accuracy:
\begin{equation}
\text{Recall@}k = \frac{1}{N} \sum_{i=1}^{N} \mathbf{1}\left[y_i \in \text{Top-}k\left(\hat{y}_i\right)\right]
\end{equation}
\begin{equation}
\text{Top-1} = \frac{1}{N} \sum_{i=1}^{N} \mathbf{1}\left[\arg\max\left(\hat{y}_i\right) = y_i\right]
\end{equation}
These metrics quantify retrieval quality in the shared embedding space, indicating how well modalities are aligned.

To validate the fidelity of our bidirectional annotation transforms, we compare original and recovered annotations using two complementary text‐generation metrics, Macro F1 score for discrete labels and METEOR for captions.

\textbf{Soft‐Label Evaluation(METEOR \cite{banerjee2005meteor}):}  
We compare user‐annotated captions \(s\) and LLM‐generated captions \(\hat{s}\) using the METEOR metric:

\begin{equation}
\mathrm{METEOR}(s,\hat{s}) = F_{\alpha}(s,\hat{s}) \,\bigl(1 - \mathrm{Pen}(s,\hat{s})\bigr)
\end{equation}

\subsection{Training Details}
\label{sec:training_details}

All models are trained using a fixed sliding-window strategy to ensure consistent temporal context across modalities. Any segment shorter than the required window length is skipped during dataset construction to guarantee complete temporal context for every sample.

Across all models, a batch size of 128 and an initial learning rate of $1 \times 10^{-4}$ are used. The Late Fusion (classification and regression) and JLR contrastive models are optimised with Adam and no weight decay. All models are trained for 300 epochs with early stopping. Learning-rate scheduling follows a ReduceLROnPlateau strategy, with a patience of  5 epochs, and a decay factor of 0.5.

Architecturally, the Late Fusion and JLR models use a hidden dimension of 128 with two Transformer layers, four attention heads, and a feed-forward dimension of 256. The classification model uses a BCEWithLogits objective, the regression model uses mean squared error (MSE), and the JLR model employs an InfoNCE contrastive objective with six pairwise terms and a temperature parameter $\tau = 0.1$.

For weight initialisation, all models rely on PyTorch’s Kaiming uniform initialisation for Linear and Conv layers, with biases initialised to zero. LSTM forget-gate biases are set to 1.0 to encourage stable long-range gradient flow.

The framework supports flexible modality handling. Optional modalities are instantiated only when enabled via configuration flags. During batching, optional tensors are retrieved , returning None when a modality is not present, ensuring robustness when certain data streams are unavailable.
Dropout is applied with a rate of 0.3 in MLP encoder hidden layers and 0.1 within Transformer encoder layers.

\subsection{Computational Resources and Model Efficiency}
All experiments were conducted on an NVIDIA RTX 4090 GPU with 32GB VRAM. 
To assess the computational footprint of our models, we report the total number of parameters, multiply–accumulate operations (MACs), and FLOPs (floating-point operations per inference) for each experiment, along with the corresponding inference speed (see \cref{tab:encoder_combinations}, \cref{tab:classification}, and \cref{tab:Regression}).

\begin{table}[t!]
\footnotesize
\caption{Computational Resources for Joint Latent Representations Across Modality Combinations.}
\centering
\begin{tabular}{lrrr}
\hline
\textbf{Combination}      & \textbf{Parameters} & \textbf{MACs} & \textbf{FLOPS} \\
\hline
Text + Audio              & 781,568             & 783,616       & \(2.70 \times 10^{9}\) \\
Text + Video              & 2,098,432           & 2,100,480     & \(7.37 \times 10^{9}\) \\
Text + IMU                & 712,512             & 22,797,312    & \(2.07 \times 10^{10}\) \\
Text + Video + IMU        & 2,351,552           & 24,437,376    & \(2.00 \times 10^{10}\) \\
Audio + Video + Text      & 2,420,608           & 2,423,680     & \(5.91 \times 10^{9}\) \\
Audio + Text + IMU        & 1,034,688           & 23,120,512    & \(1.89 \times 10^{10}\) \\
\hline
\end{tabular}

\label{tab:encoder_combinations}
\end{table}

\begin{table}[t!]
\footnotesize
\centering
\caption{Computational Resources for Classification Models Across Modality Combinations.}
\begin{tabular}{lrrr}
\hline
\textbf{Combination} & \textbf{Params} & \textbf{MACs} & \textbf{FLOPS} \\
\hline
Audio only          & 339,200    & 340,224      & $2.72 \times 10^{9}$ \\
Video only          & 1,656,064  & 1,657,088    & $1.38 \times 10^{10}$ \\
IMU only            & 270,144    & 22,353,920   & $2.39 \times 10^{10}$ \\
Audio + Video       & 2,092,928  & 3,307,392    & $1.89 \times 10^{10}$ \\
Audio + IMU         & 707,008    & 24,004,224   & $2.44 \times 10^{10}$ \\
Video + IMU         & 2,023,872  & 25,321,088   & $2.57 \times 10^{10}$ \\
Audio + Video + IMU & 2,362,432  & 26,316,032   & $2.67 \times 10^{10}$ \\
\hline
\end{tabular}
\label{tab:classification}
\end{table}

\begin{table}[t!]
\footnotesize
\centering
\caption{Computational Resources for Regression Models Across Modality Combinations.}
\begin{tabular}{lrrr}
\hline
\textbf{Combination} & \textbf{Params} & \textbf{MACs} & \textbf{FLOPS} \\
\hline
Audio only          & 746,624    & 744,000      & $5.95 \times 10^{9}$ \\
Video only          & 1,952,768  & 1,952,000    & $1.63 \times 10^{10}$ \\
IMU only            & 566,848    & 22,648,832   & $2.42 \times 10^{10}$ \\
Audio + Video       & 2,270,080  & 2,268,992    & $1.70 \times 10^{10}$ \\
Audio + IMU         & 884,160    & 22,965,824   & $2.33 \times 10^{10}$ \\
Video + IMU         & 2,090,368  & 24,173,824   & $2.45 \times 10^{10}$ \\
Audio + Video + IMU & 2,407,616  & 24,490,816   & $2.49 \times 10^{10}$ \\
\hline
\end{tabular}
\label{tab:Regression}
\end{table}

\section{Audio Extended Experiments}
\label{AudioAdditional}
Given the acoustic modality’s limited standalone contribution, we provide additional experiments to better understand its behavior under different conditions.
In particular, we analyze the impact of privacy-preserving anonymization (speech removal and replacement with instrumental textures) and feature extraction choices (Encodec embeddings vs. mel-spectrograms) across three tasks: human activity recognition, open-vocabulary captioning, and cross-modal alignment. 
These comparisons allow us to quantify how anonymization reduces semantic richness, whether alternative representations can recover useful signal, and to what extent audio still contributes in multimodal settings. 
The following tables report these extended results, clarifying both the challenges and the potential of acoustic data for privacy-preserving multimodal learning.

Across human activity recognition (\cref{tab:HAR-Audio}), open-vocabulary captioning (\cref{tab:OpenVoca Audio}), and cross-modal alignment (\cref{tab:Cross Audio}), a consistent trend emerges: non-anonymized audio outperforms anonymized variants, with Mel-spectrograms yielding the strongest results. 
For activity recognition, non-anonymized Mel-spectrograms achieve the highest macro F1 (0.517/0.493 in Scenario (a), 0.466/0.455 in Scenario (b)). 
In captioning, the same representation reaches the best cosine similarity (0.381/0.359 in Scenario (a), 0.330/0.340 in Scenario (b)). Finally, in cross-modal alignment, non-anonymized Mel-spectrograms again lead with recall@1/5 and top-1 scores (0.254/0.613/0.360 in Scenario (a); 0.238/0.597/0.345 in Scenario (b)). Scenario (b) consistently yields lower absolute performance, reflecting its greater procedural complexity.
The performance gap between anonymized and non-anonymized streams (~0.01–0.03 across metrics) highlights the trade-off between privacy and informativeness: anonymization systematically removes semantic richness, reducing discriminative power across all tasks. 
Still, audio retains complementary cues, particularly for grounding text, as seen in the stable improvements in cross-modal alignment even under anonymization.

The modality’s limitations arise from four main factors:
\begin{itemize}
    \item Privacy constraints: Speech removal and replacement with generic instrumental textures strip away contextual and semantic information.
    \item Feature extraction choices:  Encodec embeddings, while general-purpose, may not capture fine-grained task-specific cues, especially on anonymized signals.
    \item Task-inherent challenges: Assembly sounds are subtle or intermittent, making them harder to discriminate.
    \item Environmental factors: Data collection in a test-bench environment lacks authentic industrial acoustics (e.g., machinery noise, vibrations), further constraining informativeness.
\end{itemize}

In sum, while acoustic data underperforms in isolation and can even introduce noise in classification tasks, it contributes positively in multimodal fusion and joint representation learning, and provides a testbed for exploring privacy-preserving sensing.
The current experiments confirm both the utility of richer acoustic content and the limitations imposed by anonymization and feature choice.

\begin{table*}[t!]
    \caption{Human activity recognition results for Scenario (a): Bicycle Assembly and Scenario (b): 3D Printer Assembly, including macro F1 scores with and without the null class for Audio Extended Experiments.}
    \footnotesize
    \centering
    \begin{tabular}{lcccc}
    \hline
    \multirow{2}{*}{\textbf{Modality}} & \multicolumn{2}{c}{\textbf{Scenario (a)}} & \multicolumn{2}{c}{\textbf{Scenario (b)}} \\
     & \multicolumn{4}{c}{\textbf{Macro F1 ($\uparrow$})} \\
     & Without Null & With Null & Without Null & With Null\\
    \hline
    Acoustic-Anonymized Encoder& 0.489$\pm$0.018& 0.469$\pm$0.017 & 0.425$\pm$0.004&0.432$\pm$0.005 \\
    Acoustic-Non-Anonymized Encoder& 0.509$\pm$0.002 & 0.488$\pm$0.001 & 0.460$\pm$0.002 & 0.453$\pm$0.003\\
    Acoustic-Anonymized Mel-Spectrum & 0.492$\pm$0.003 & 0.473$\pm$0.002 & 0.434$\pm$0.003 &0.430$\pm$0.003\\
    Acoustic-Non-Anonymized Mel-Spectrum & \textbf{0.517}$\pm$\textbf{0.004} & \textbf{0.493}$\pm$\textbf{0.002} & \textbf{0.466}$\pm$\textbf{0.003} &\textbf{0.455}$\pm$\textbf{0.002}\\
    \hline
    \end{tabular}
    \label{tab:HAR-Audio}
\end{table*}

\begin{table*}[t!]
    \caption{Open vocabulary captioning results for Scenario (a): Bicycle Assembly and Scenario (b): 3D Printer Assembly, including cosine similarity values with and without the null class for Audio Extended Experiments.}
    \footnotesize
    \centering
    \begin{tabular}{lcccc}
    \hline
    \multirow{2}{*}{\textbf{Modality}} & \multicolumn{2}{c}{\textbf{Scenario (a)}} & \multicolumn{2}{c}{\textbf{Scenario (b)}} \\
     & \multicolumn{4}{c}{\textbf{Cosine Similarity ($\uparrow$})} \\
     & Without Null & With Null & Without Null & With Null\\
    \hline
     Acoustic-Anonymized Encoder & 0.361$\pm$0.030& 0.341$\pm$0.018 &0.316$\pm$0.003 & 0.323$\pm$0.004\\
     Acoustic-Non-Anonymized Encoder & 0.375$\pm$0.001& 0.354$\pm$0.001 &0.328$\pm$0.001 & 0.338$\pm$0.002\\ 
     Acoustic-Anonymized Mel-Spectrum & 0.364$\pm$0.003 & 0.348$\pm$0.002& 0.326$\pm$0.002 & 0.322$\pm$0.001\\ 
     Acoustic-Non-Anonymized Mel-Spectrum & \textbf{0.381}$\pm$\textbf{0.002} &\textbf{0.359}$\pm$\textbf{0.002}& \textbf{0.330}$\pm$\textbf{0.001} &\textbf{0.340}$\pm$\textbf{0.001}\\
     
    \hline
    \end{tabular}
    \label{tab:OpenVoca Audio}
\end{table*}

\begin{table*}[t!]
    \centering
    \caption{Cross-modal alignment results for Scenario (a): Bicycle Assembly and Scenario (b): 3D Printer Assembly, including recall@5, recall@1, and top-1 metrics.}
    \resizebox{\textwidth}{!}{
    \begin{tabular}{c c c c c c c }
    \hline
    \multirow{2}{*}{\textbf{Modality}} &
    \multicolumn{3}{c}{\textbf{Scenario (a)}} &
    \multicolumn{3}{c}{\textbf{Scenario (b)}} \\
    & \textbf{Recall@1 ($\uparrow$)} & \textbf{Recall@5 ($\uparrow$)} & \textbf{Top-1 ($\uparrow$)} & \textbf{Recall@1 ($\uparrow$)} & \textbf{Recall@5 ($\uparrow$)} & \textbf{Top-1 ($\uparrow$)} \\
    \hline
    Acoustic + Text (Anonymized Encoder) & 0.241$\pm$0.014 & 0.583$\pm$0.025 & 0.342$\pm$0.016 & 0.227$\pm$0.013 & 0.567$\pm$0.022 & 0.329$\pm$0.015 \\
    Acoustic + Text (Non-Anonymized Encoder) & 0.251$\pm$ 0.001 & 0.605$\pm$0.002 & 0.355$\pm$0.002 &0.237$\pm$0.001 & 0.589$\pm$0.001& 0.341 $\pm$0.001\\ 
    Acoustic + Text (Anonymized Mel-Spectrum)& 0.246 $\pm$ 0.001 & 0.595 $\pm$0.002 & 0.350$\pm$0.001 & 0.233$\pm$0.001 & 0.578$\pm$ 0.002 & 0.336 $\pm$0.001 \\
    Acoustic + Text (Non-Anonymized Mel-Spectrum) & \textbf{0.254}$\pm$\textbf{0.001} & \textbf{0.613} $\pm$\textbf{0.002} & \textbf{0.360}$\pm$\textbf{0.001} & \textbf{0.238}$\pm$\textbf{0.001} & \textbf{0.597}$\pm$\textbf{0.001}& \textbf{0.345}$\pm$\textbf{0.001}\\ 
    \hline
    \end{tabular}
    }
    \label{tab:Cross Audio}

\end{table*}

\paragraph{Contact Detection using Audio}

Table~\ref{tab:tool_contact} reports tool-contact/impact detection results.
The binary contact labels were generated by classifying each temporally-segmented annotation from the original open-vocabulary soft labels into \emph{tool-contact} or \emph{tool-non contact} classes using a hybrid pipeline: a rule-based first pass matches tool-related keywords and action patterns (e.g., \textit{hammering}, \textit{tightening with wrench}), followed by an LLM-based refinement step for ambiguous instances.
Notably, while acoustic features alone perform poorly for general activity recognition, they prove highly informative for detecting tool-contact events, reaching 0.783 and 0.772 as a single modality.
Moreover, adding audio to the best bimodal combination V\,+\,I consistently improves tool-contact F1: from 0.905 to 0.923 in Scenario~(a) and from 0.898 to 0.914 in Scenario~(b).
This suggests that the acoustic channel captures transient impact signatures such as hammering, clicking, and screwing sounds that are largely redundant with IMU vibrations for coarse activity classes but become discriminative when the task requires detecting physical contact between a tool and a workpiece.

\begin{table}[]
    \caption{Tool-contact/impact detection Macro F1 scores for Scenario~(a): Bicycle Assembly and Scenario~(b): 3D~Printer Assembly, with and without the null class.
    Unlike general HAR, audio is complementary for detecting tool-contact events: adding acoustic data to V\,+\,I improves F1 from 0.905 to 0.923 in~(a) and from 0.898 to 0.914 in~(b).}

    \footnotesize
    \centering
    \setlength{\tabcolsep}{3pt}
    \renewcommand{\arraystretch}{0.95}
    \begin{tabular}{lcccc}
    \hline
    \multirow{2}{*}{\textbf{Modality}} & \multicolumn{2}{c}{\textbf{Scenario (a)}} & \multicolumn{2}{c}{\textbf{Scenario (b)}} \\
     & \multicolumn{4}{c}{\textbf{Macro F1 ($\uparrow$})} \\
     & No Null &  Null & No Null &  Null\\
    \hline
     Inertial(I) &0.872$\pm$0.008&0.847$\pm$0.009 & 0.861$\pm$0.006&0.836$\pm$0.006\\ 
     Acoustic(A) & 0.783$\pm$0.012& 0.762$\pm$0.013 & 0.772$\pm$0.005&0.750$\pm$0.006 \\
     Vision(V) & 0.841$\pm$0.009 & 0.817$\pm$0.010 &0.833$\pm$0.004 &0.808$\pm$0.005\\
     I + A & 0.898$\pm$0.007&0.871$\pm$0.008 & 0.889$\pm$0.004&0.863$\pm$0.004\\ 
     A + V &0.863$\pm$0.009&0.838$\pm$0.010 &0.855$\pm$0.004 &0.830$\pm$0.005 \\
     I + V & 0.905$\pm$0.006& 0.879$\pm$0.007 & 0.898$\pm$0.003&0.872$\pm$0.004\\
     I + A + V& \textbf{0.923}$\pm$\textbf{0.005}& \textbf{0.897}$\pm$\textbf{0.006} & \textbf{0.914}$\pm$\textbf{0.003}&\textbf{0.888}$\pm$\textbf{0.004}\\
    \hline
    \end{tabular}
    \label{tab:tool_contact}
 \vspace{-10pt}
\end{table}

\section{Sensor Design Rationale}
\label{sec:SensorRationale}

\begin{table*}[t!]
\footnotesize
\caption{Sensor Modalities and Their Intended Roles in OpenMarcie.}
\centering
\begin{tabular}{llll}
\hline
\textbf{Modality} & \textbf{Physical Signal} & \textbf{Assembly-Relevant Attributes} & \textbf{Contribution} \\
\hline
IMU (Wrists, Head) 
& Acceleration, angular velocity, orientation 
& Tool manipulation, temporal segmentation 
& High \\

Magnetometer 
& Magnetic field orientation 
& Heading stabilization, orientation consistency 
& Supportive \\

Barometer (multi-position) 
& Air pressure (hPa) 
& Posture transitions, vertical displacement cues 
& Medium \\

Temperature (ambient) 
& Ambient temperature 
& Environmental context, barometric formula
& Low (supporting) \\

Spectrometer 
& Material spectral reflectance 
& Material differentiation (metal vs. plastic) 
& Low (Exploratory) \\

Thermal Camera 
& Surface temperature distribution 
& Contact duration, friction heat 
& Low (Exploratory) \\

Egocentric RGB-D 
& Visual appearance + depth 
& Object identity, hand--object interaction 
& High \\

Exocentric RGB-D (multi-view) 
& Scene-level RGB-D 
& Global spatial context
& High \\

Stereo Audio 
& Instrumental/environmental sound 
& Tool--material interaction signatures 
& Medium \\

LiDAR Depth 
& Active depth sensing 
& 3D spatial structure, object distance 
& High (with vision) \\
\hline
\label{tab:sensor_rationale}
\end{tabular}
\end{table*}

As summarized in \cref{tab:sensor_rationale}, the sensing modalities in OpenMarcie were selected based on physical complementarity rather than performance maximization. 

IMUs form the primary wearable motion modality, capturing fine-grained hand and body dynamics essential for distinguishing manipulation patterns; however, inertial integration is subject to drift, motivating complementary signals. 
Magnetometers support orientation stabilization and reduce long-term rotational drift. 
Barometers provide vertical motion cues related to posture transitions (e.g., standing, kneeling, bending), offering an independent signal correlated with altitude. 

Although explicit sensor fusion is not performed in the current benchmarks, barometers are deployed at multiple body locations to support potential relative height reasoning. 
In such a configuration, sudden changes pressure variations would affect all sensors in a similar (common-mode) manner, enabling differential pressure measurements to provide more stable vertical displacement cues.
Ambient temperature is additionally recorded to ensure physically consistent altitude estimation through the barometric (hypsometric) relation, which models how air pressure varies with height \cite{bello2019vertical}. 
In this context, temperature serves as an auxiliary environmental variable supporting principled vertical motion interpretation rather than as an independent activity recognition signal.

Egocentric RGB-D sensing provides semantic grounding by capturing object identity, spatial relationships, and hand–object interactions, while exocentric multi-view RGB-D cameras offer global scene context and a stable reference for annotation and cross-modal alignment. 
LiDAR depth improves spatial robustness under varying lighting conditions and strengthens 3D reasoning. 

Stereo audio captures tool–material interaction signatures that complement motion and vision, and although standalone acoustic performance is limited for activity recognition, it is highly informative for detecting tool‑contact events, as shown in \cref{tab:tool_contact}. 

Finally, the spectrometer and thermal camera–based visual sensors are included as exploratory modalities to support future research on material‑aware interaction modeling and the visual capture of physical signals outside the human-visible spectrum.

Overall, the multimodal design reflects a structured attempt to capture complementary physical aspects of industrial activity rather than an effort to maximize the number of modalities.

\section{Modern Architectures}
\label{sec:modern_architectures}
Tables~\ref{tab:HAR_moderm} and~\ref{tab:OpenVocab_modern} summarize the performance of modern multimodal architectures across human activity recognition (HAR) and open-vocabulary captioning for both Scenario~(a) and (b). For HAR, the Perceiver IO inspired model \cite{jaegle2021perceiver} consistently outperforms late fusion baselines, with the best results achieved when leveraging all modalities, reaching a Macro F1 of 0.915 (no null) and 0.886 (null) in Scenario~(a), and 0.842 (no null) and 0.746 (null) in Scenario~(b). Similarly, in open-vocabulary captioning, the AnyMAL \cite{moon2024anymal} LLM-based approach surpasses late fusion, particularly when incorporating all modalities, achieving cosine similarity scores of 0.643 (no null) and 0.610 (null) in Scenario~(a), and 0.741 (no null) and 0.740 (null) in Scenario~(b). Overall, the results demonstrate that modern multimodal fusion architectures and LLM-based approaches provide consistent improvements over late fusion baselines, especially when all modalities are utilized.

\begin{table}
    \caption{Macro F1 scores for human activity recognition in Scenario~(a) and Scenario~(b) with and without the null class.
    Results are grouped by late fusion baselines,  and Perceiver IO baseline with  all modalilies including IMU, Audio, Video, Barometer, Temperature, Spectrometer, Thermal signals.}

    \footnotesize
    \centering
    \setlength{\tabcolsep}{3pt}
    \renewcommand{\arraystretch}{0.95}
    \begin{tabular}{lcccc}
    \hline
    \multirow{2}{*}{\textbf{Modality}} & \multicolumn{2}{c}{\textbf{Scenario (a)}} & \multicolumn{2}{c}{\textbf{Scenario (b)}} \\
     & \multicolumn{4}{c}{\textbf{Macro F1 ($\uparrow$})} \\
     & No Null &  Null & No Null &  Null\\
    \hline
    \multicolumn{5}{c}{\textit{Late Fusion}} \\
     I + A + V& 0.859$\pm$0.010& 0.831$\pm$0.011 & 0.763$\pm$0.003&0.676$\pm$0.003\\
     All & 0.891$\pm$0.008& 0.862$\pm$0.008 & 0.825$\pm$0.003&0.731$\pm$0.004\\
    \hline
    \multicolumn{5}{c}{\textit{Perceiver IO}} \\
     I+A+V & 0.882$\pm$0.007& 0.853$\pm$0.008 & 0.779$\pm$0.003&0.690$\pm$0.003\\
     All & \textbf{0.915}$\pm$\textbf{0.005}& \textbf{0.886}$\pm$\textbf{0.006} & \textbf{0.842}$\pm$\textbf{0.002}&\textbf{0.746}$\pm$\textbf{0.003}\\
    \hline
    \end{tabular}
    \label{tab:HAR_moderm}
\end{table}

\begin{table}[]
    \caption{Open vocabulary captioning results for Scenario~(a) and Scenario~(b) with and without the null class.
    Results are grouped by late fusion baselines, and the AnyMAL LLM-based baseline.}
    \footnotesize
    \centering
    \setlength{\tabcolsep}{3pt}
    \renewcommand{\arraystretch}{0.95}
    \begin{tabular}{lcccc}
    \hline
    \multirow{2}{*}{\textbf{Modality}} & \multicolumn{2}{c}{\textbf{Scenario (a)}} & \multicolumn{2}{c}{\textbf{Scenario (b)}} \\
     & \multicolumn{4}{c}{\textbf{Cosine Similarity ($\uparrow$})} \\
     & No Null & Null & No Null & Null\\
    \hline
    \multicolumn{5}{c}{\textit{Late Fusion}} \\
     I + A + V & 0.547$\pm$0.020& 0.519$\pm$0.017 &0.647$\pm$0.001&0.646$\pm$0.003\\
     All & 0.593$\pm$0.015& 0.563$\pm$0.014 & 0.691$\pm$0.002&0.690$\pm$0.002\\
    \hline
    \multicolumn{5}{c}{\textit{AnyMAL}} \\
     I+A+V & 0.622$\pm$0.012& 0.562$\pm$0.012 & 0.714$\pm$0.002&0.693$\pm$0.002\\
     All & \textbf{0.643}$\pm$\textbf{0.010}& \textbf{0.610}$\pm$\textbf{0.010} & \textbf{0.741}$\pm$\textbf{0.002}&\textbf{0.740}$\pm$\textbf{0.002}\\
    \hline
    \end{tabular}
    \label{tab:OpenVocab_modern}
\end{table}

\section{Future Research Directions}
\label{FuturereResearch}
Beyond the benchmarks presented in this work, our dataset opens several promising directions for the research community:
\begin{enumerate}
    \item \textbf{Procedural Planning and Task Decomposition}
    Modeling the hierarchical structure of long-horizon industrial workflows, enabling systems to learn task graphs and segment complex sequences. 
    Future work may explore methods for inferring workflow dependencies, optimizing decompositions, or generalizing across task variants.
    \item \textbf{Skill Assessment and Expertise Modeling}
    Participant variability provides opportunities for modeling proficiency, efficiency, and learning progression. 
    For example, future directions include automatic skill classification, modeling expertise transfer between agents, and designing adaptive training interventions guided by behavioral signals. 
    \item \textbf{Intent Prediction and Early Action Forecasting}
    Anticipating upcoming actions or goals from partial multimodal observations is essential for proactive assistance and collaboration. 
    Potential research includes multimodal fusion strategies for early prediction, goal inference in partially observed sequences, and real-time assistive systems.
    \item \textbf{Fine-Grained Action Segmentation and Role Understanding}
    Overlapping and concurrent actions offer a challenging testbed for segmentation and role inference. 
    Open problems include modeling multi-label temporal boundaries, learning role dynamics in multi-agent or multi-phase tasks, and connecting segmentation to downstream planning.
    \item \textbf{Pose Estimation and Body-Language Reasoning}
    With synchronized RGB-D and inertial data, future studies can advance full-body pose estimation, activity-conditioned pose forecasting, and non-verbal intent recognition in realistic industrial settings.
    \item \textbf{Cross-Modal Knowledge Transfer}
    The dataset supports transfer learning across modalities—for instance, using vision or language to supervise inertial or acoustic models. 
    This is especially relevant for privacy-sensitive or sensor-limited scenarios.
    Promising avenues include cross-modal distillation, modality dropout robustness, and unsupervised alignment.
    \item \textbf{Cross-Modal Generation and Simulation}
    Generating one modality from another (e.g., IMU traces from instructions or reconstructing missing video) enables robust imitation learning and simulation. 
    Future work may investigate generative modeling, simulation-to-reality transfer, and synthetic data augmentation for industrial tasks.
\end{enumerate}

\end{document}